%% file: _main.tex
\input{_constants}
\arxiv

\pdfoutput=1
\documentclass[10pt,twocolumn,letterpaper]{article}
\input{cvpr_header}
\begin{document}
\title{\paperTitle}
\author{\authorBlock}

\maketitle

\input{00_abstract}
\input{01_intro}
\input{02_related}

\input{04_method}
\input{08_experiments}

\input{10_conclusion}

{\small
\bibliographystyle{ieeenat_fullname}
\bibliography{11_references}
}

\ifarxiv \clearpage \appendix \input{12_appendix} \fi

\end{document}

%% file: _constants.tex
\def\paperID{3739}
\def\confName{CVPR}
\def\confYear{2024}

\def\paperTitle{No More Ambiguity in 360$^\circ$ Room Layout via Bi-Layout Estimation} 

\def\authorBlock{
    Yu-Ju Tsai$^{1,3}$\thanks{Equal contribution}
    \qquad
    Jin-Cheng Jhang$^{2}$\footnotemark[1] \qquad
    Jingjing Zheng$^{3}$ \qquad
    Wei Wang$^{3}$\\
    Albert Y. C. Chen$^{3}$ \qquad
    Min Sun$^{2,3}$ \qquad
    Cheng-Hao Kuo$^{3}$ \qquad
    Ming-Hsuan Yang$^{1}$\\
    $^{1}$UC Merced \quad 
    $^{2}$National Tsing Hua University \quad 
    $^{3}$Amazon\\
    {\tt\small \{ytsai17, myang37\}@ucmerced.edu \quad
    \{frank890725\}@gapp.nthu.edu.tw}
    \\
    {\tt\small \{zhejingj, wweiwan, aycchen, minnsun, chkuo\}@amazon.com
    }
}

\newif\ifreview 
\newif\ifarxiv \newcommand{\arxiv}{\arxivtrue}
\newif\ifcamera 
\newif\ifrebuttal 

%% file: cvpr_header.tex
\ifreview \usepackage[review]{cvpr} \fi
\ifarxiv \usepackage[pagenumbers]{cvpr} \fi
\ifrebuttal \usepackage[rebuttal]{cvpr} \fi
\ifcamera \usepackage{cvpr} \fi

\input{_macros}  

\usepackage{xr-hyper}

\makeatletter
\newcommand*{\addFileDependency}[1]{
  \typeout{(#1)}
  \@addtofilelist{#1}
  \IfFileExists{#1}{}{\typeout{No file #1.}}
}

\makeatother

\definecolor{ourcolor}{rgb}{0.95, 0.90, 0.95}
\definecolor{cvprblue}{rgb}{0.21,0.49,0.74}
\usepackage[pagebackref,breaklinks,colorlinks,citecolor=cvprblue,linkcolor=cvprblue,urlcolor=cvprblue]{hyperref}
\usepackage[capitalize]{cleveref}
\crefname{section}{Sec.}{Secs.}
\crefname{table}{Table}{Tables}
\crefname{figure}{Fig.}{Figs.}

\newcommand{\red}[1]{{\textcolor{red}{#1}}} 
\newcommand{\blue}[1]{{\textcolor{blue}{#1}}} 
\newcommand{\green}[1]{{\textcolor{green}{#1}}} 


%% file: _macros.tex
\usepackage[accsupp]{axessibility} 

\usepackage{graphicx}	
\usepackage{amsmath}	
\usepackage{amssymb}	
\usepackage{booktabs}
\usepackage{times}
\usepackage{microtype}
\usepackage{epsfig}
\usepackage[table,xcdraw,dvipsnames]{xcolor}
\usepackage{caption}
\usepackage{float}
\usepackage{placeins}
\usepackage{color, colortbl}
\usepackage{stfloats}
\usepackage{enumitem}
\usepackage{tabularx}
\usepackage{xstring}
\usepackage{multirow}
\usepackage{xspace}
\usepackage{url}
\usepackage{subcaption}
\usepackage{xcolor}
\usepackage{comment}
\usepackage[hang,flushmargin]{footmisc}
\DeclareUnicodeCharacter{2218}{\ensuremath{\circ}}

\ifcamera \usepackage[accsupp]{axessibility} \fi

% ...

\ifarxiv  \fi

\newcommand{\R}[1]{{%
    \textbf{%
        \ifstrequal{#1}{1}{\textcolor{red}{R#1}}{%
        \ifstrequal{#1}{2}{\textcolor{blue}{R#1}}{%
        \ifstrequal{#1}{3}{\textcolor{magenta}{R#1}}{%
        \ifstrequal{#1}{4}{\textcolor{teal}{R#1}}{%
                           \textcolor{cyan}{R#1}%
        }}}}%
    }%
}}

%% file: 00_abstract.tex
\begin{abstract}
Inherent ambiguity in layout annotations poses significant challenges to developing accurate 360$^\circ$ room layout estimation models. 
To address this issue, we propose a novel Bi-Layout model capable of predicting two distinct layout types. 
One stops at ambiguous regions, while the other extends to encompass all visible areas. 
Our model employs two global context embeddings, where each embedding is designed to capture specific contextual information for each layout type. 
With our novel feature guidance module, the image feature retrieves relevant context from these embeddings, generating layout-aware features for precise bi-layout predictions.
A unique property of our Bi-Layout model is its ability to inherently detect ambiguous regions by comparing the two predictions. To circumvent the need for manual correction of ambiguous annotations during testing, we also introduce a new metric for disambiguating ground truth layouts. 
Our method demonstrates superior performance on benchmark datasets, notably outperforming leading approaches.
Specifically, on the MatterportLayout dataset, it improves 3DIoU from $81.70\%$ to $82.57\%$ across the full test set and notably from $54.80\%$ to $59.97\%$ in subsets with significant ambiguity. 

\end{abstract}

%% file: 01_intro.tex
\section{Introduction}
\label{sec:intro}
Room layout estimation from a single 360$^\circ$ image has received significant attention due to the availability of cheap 360$^\circ$ cameras and the demonstration of visually pleasing room pop-ups. It also plays a vital role in indoor 3D scene understanding~\cite{sun2021hohonet,huang2018holistic,zhang2021holistic,chen2022pq} as the room layout constrains the space where objects are placed and interact. 
Its performance has improved significantly over the years, where the gain comes from better algorithms design~\cite{sun2019horizonnet,wang2021led2,jiang2022lgt,shen2023dopnet}, and more challenging data collected~\cite{zou2021manhattan,cruz2021zillow}. Despite the progress, the task formulation of predicting a \textit{single} layout given a \textit{single} 360$^\circ$ image has never been changed.


\input{figs/motivation}

Annotating a \textit{single} layout for each 360$^\circ$ image is, in fact, an ambiguous task. For instance, consider the two images with openings in Fig.~\ref{fig:motivation}, where the ground truth (GT) annotation stops at openings and encloses the nearest room in Fig.~\ref{fig:motivation}(a) while extending to all visible areas inside the opening in Fig.~\ref{fig:motivation}(b). 
Notably, even within the same dataset, there are variations in how opening regions are annotated. We observe that this ambiguity issue in annotation is prevalent across most datasets, and there is a lack of a clear definition of how to annotate ambiguous regions. 
Furthermore, state-of-the-art (SoTA) methods often overlook this ambiguity issue, leading to inherent inaccuracy during training. 
As a result, existing methods may predict in a manner opposite to the GT, as shown in Fig.~\ref{fig:motivation}. 
In this paper, we define two main types of layout annotations: \textbf{\textit{enclosed}} and \textbf{\textit{extended}}. The former stops at ambiguous regions and encloses the room, whereas the latter extends to all visible areas inside the opening. 

To address the confusion arising from the ambiguity issue in model training, as shown in Fig.~\ref{fig:network}, we propose a novel Bi-Layout model to simultaneously predict both \textit{enclosed} and \textit{extended} layouts for each image. Our model consists of three components: a shared feature extractor, two separate global context embeddings, and a shared feature guidance module. Two separate global context embeddings are learned to encode all context information related to the corresponding layout type. The shared feature guidance module guides the fusion of the shared image feature with our two embeddings separately through cross-attention. Specifically, we use the image feature as the query and each global context embedding as the key and value. When queried by the image feature, the global context embedding can inject layout type-related context information into the image feature. This results in enhanced alignment of the image feature with the corresponding layout prediction type. 
\input{figs/comparison}

Our model design introduces two key innovations. First, contrary to the standard Query-Key-Value setting employed in DETR~\cite{carion2020end} and other high-level tasks~\cite{cheng2021per,cheng2022masked,meinhardt2022trackformer,zhou2022global,yue2023connecting,su2023slibo}, where embeddings serve as queries to retrieve relevant information from the image feature, we invert this relationship and employ our image feature as the queries. This unconventional design allows the image feature to be guided by our embedding, which represents the global information of a specific layout type. To the best of our knowledge, we are the first to develop a query-based model for room layout estimation, inherently designed for predicting multiple layouts. The second innovation lies in the efficiency of our model, which can predict two layouts with minimal additional model size overhead. For bi-layout estimations, naive approaches either train two distinct models with identical architectures on different labels or train a single model by sharing the feature extractor but separating other components. However, the former method doubles the model size and training time, while the latter lacks compactness and grapples with interference in simultaneously learning two layout types. In contrast, our model is not only the smallest, achieved by sharing both the feature extractor and the guidance module, but it also avoids interference issues by employing separate global context embeddings to guide feature fusion for different layout types. As shown in Fig.~\ref{fig:ambiguity_results}, our model is able to predict two extremely different layouts.

We also introduce a new metric termed as \textit{disambiguate} metric to resolve ambiguities in the annotations of test data. It calculates the Intersection over Unions (IoU) of both predicted layouts with the ground truth and selects the higher IoU for evaluation. This is an effective way to quantitatively measure the benefit of our Bi-Layout estimation without manually correcting ambiguous annotations during testing. Another noteworthy feature of our Bi-Layout model is its ability to detect ambiguous regions with reasonable precision and recall by comparing two predictions. 
Our method exhibits superior performance on benchmark datasets, surpassing SoTA methods. On MatterportLayout~\cite{zou2021manhattan}, it enhances 3DIoU from $81.70\%$ to $82.57\%$ on the entire test set and notably from $54.80\%$ to $59.97\%$ in subsets with substantial ambiguity. 
The main contributions of this work are:
\begin{itemize}
    \item We clearly identify layout ambiguity issues in existing datasets and propose a \textit{disambiguate} metric to measure the accuracy with multiple predictions effectively.
    \item We propose a novel Bi-Layout model that utilizes two global context embeddings with a shared feature guidance module to generate multiple layout predictions while keeping the model compact.
    \item We evaluate our method with extensive experiments and prove it outperforms SoTA methods in all settings, showing that our Bi-Layout model effectively resolves the layout ambiguity issues.
\end{itemize}
%

%% file: figs/motivation.tex
\begin{figure}[tp]
    \centering
    \includegraphics[width=\linewidth]{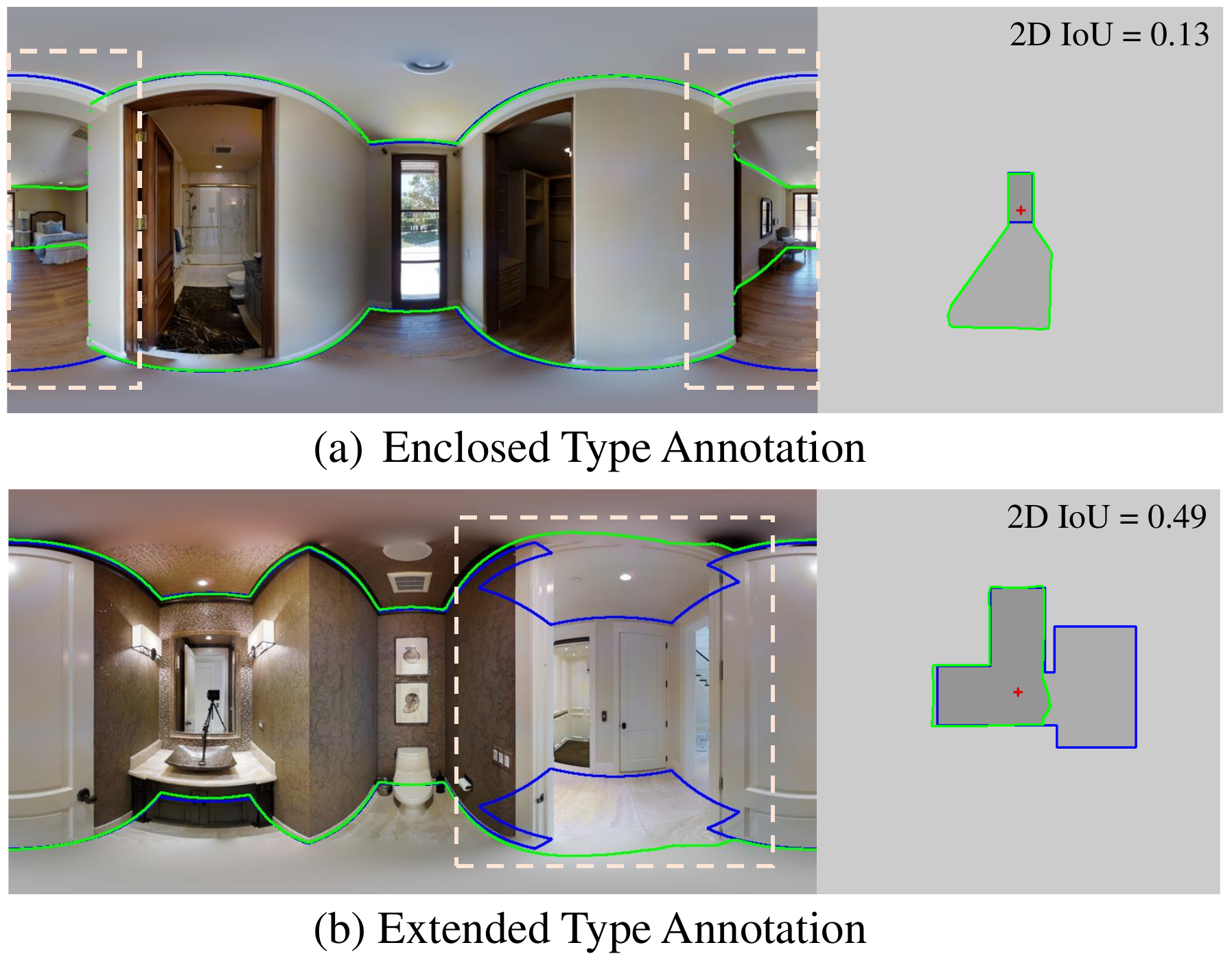}
    \vspace{-4mm}
    \caption{\textbf{Inherent ambiguity in the MatterportLayout~\cite{zou2021manhattan}.} \blue{Blue} and \green{Green} represent ground truth annotations and predictions from the SoTA models, respectively. 
    The layout boundaries are shown on the left, and their bird's-eye view projections are on the right. 
    We define two types of layout annotation: (a) \textbf{enclosed type} encloses the room. (b) \textbf{extended type} extends to all visible areas. The dashed lines underscore the ambiguity in the dataset labels.
}
    \label{fig:motivation}
    \vspace{-4mm}
\end{figure}


%% file: figs/comparison.tex
\begin{figure*}[tp]
    \centering
    \includegraphics[width=\textwidth]{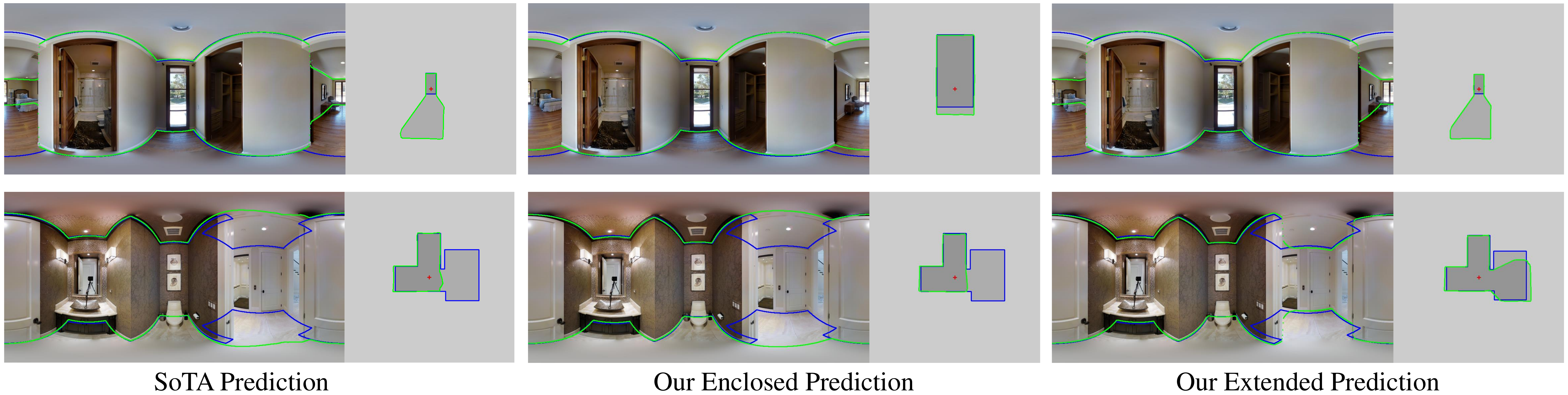}
    \vspace{-6mm}
    \caption{\textbf{Comparions of our Bi{-}Layout model and the SoTA models.} \blue{Blue} and \green{Green} indicate ground truth labels and predictions, respectively. 
    For each image, the layout boundaries are shown on the left, while their bird's-eye view projections are shown on the right. Our Bi-Layout model can predict two extremely different types of layouts (\textit{enclosed} and \textit{extended}), addressing the ambiguity issue that the SoTA methods struggle with.}
    \label{fig:ambiguity_results}
    \vspace{-4mm}
\end{figure*}

%% file: 02_related.tex
\section{Related Work}
\label{sec:related}

\vspace{-2mm}
\noindent \textbf{360$^\circ$ room layout estimation.}
In 360$^\circ$ room layout estimation, prior methods follow the Manhattan World assumption~\cite{coughlan1999manhattan}. For instance, LayoutNet~\cite{zou2018layoutnet} predicts corner and boundary probability maps directly from panoramas. Dula-Net~\cite{yang2019dula} predicts 2D-floor plane semantic masks from equirectangular and perspective views of ceilings. 
Zou~\etal presents improved versions, LayoutNet v2, and Dula-Net v2~\cite{zou2021manhattan}, demonstrating enhanced performance on cuboid datasets. 
Fernandez~\etal adopts equirectangular convolutions (EquiConvs)~\cite{fernandez2020corners} for generating corner and edge probability maps.
HorizonNet~\cite{sun2019horizonnet} and HoHoNet~\cite{sun2021hohonet} simplify layout estimation by employing 1D representations and employing Bi-LSTM~\cite{hochreiter1997long,schuster1997bidirectional} and multi-head self-attention mechanisms~\cite{vaswani2017attention} to establish long-range dependencies. LE$\mathrm{D}^{2}$-Net~\cite{wang2021led2} reformulates layout estimation as predicting the depth of walls in the horizontal direction. AtlantaNet~\cite{pintore2020atlantanet} predicts room layouts by combining projections of the floor and ceiling planes. DMH-Net~\cite{zhao20223d} transforms panorama into cubemap~\cite{greene1986environment} and predicts the position of intersection lines with learnable Hough Transform Block~\cite{zhao2021deep}. 
LGT-Net~\cite{jiang2022lgt} employs self-attention transformers~\cite{vaswani2017attention} to learn geometric relationships and capture long-range dependencies. DOP-Net~\cite{shen2023dopnet} disentangles
1D feature by segmenting features into orthogonal plane representation, and uses GCN~\cite{kipf2017semi} and transformer~\cite{vaswani2017attention} to refine the features.

These methods are designed only to predict a single layout, which often encounters challenges posed by the inherent ambiguity in dataset labels, resulting in suboptimal performance. In contrast, our Bi{-}Layout model addresses this issue by generating two distinct layout predictions through the innovative integration of global context embeddings and our shared feature guidance module design.

\input{figs/network}

\noindent \textbf{Multiple layout hypotheses.}
Several prior studies~\cite{hedau2009recovering, gupta2010estimating,schwing2012efficient,ramalingam2013manhattan,hirzer2020smart,mallya2015learning, ren2017coarse,wang2013discriminative,zhao2017physics} have employed multiple hypotheses in their methods for estimating room layouts. 
The fundamental concept behind these methods involves leveraging vanishing points, edges, or other pertinent information to generate several rays or boxes as potential layout hypotheses. 
Through various scoring function designs, one of the hypotheses can be selected as the prediction that best fits the room.
In contrast, our method generates two distinct layout predictions, which can also be viewed as having two hypotheses.
However, the key disparity lies in the fact that the previous methods only have one hypothesis defining the correct geometry. 
On the contrary, both of our predictions are meaningful and offer two different geometries, \textit{extended} and \textit{enclosed} layouts, allowing for flexibility in choosing the suitable one based on the specific requirements of different use cases.

\noindent \textbf{Query-based vision transformer.}
Transformers~\cite{vaswani2017attention} have exhibited considerable efficacy in various high-level computer vision tasks, including object detection~\cite{carion2020end,zhu2021deformable}, segmentation~\cite{cheng2021per,cheng2022masked}, tracking~\cite{meinhardt2022trackformer,zhou2022global}, and floorplan reconstruction~\cite{yue2023connecting,su2023slibo}. 
The standard transformer decoder utilizes feature embeddings as queries to extract relevant features from the image feature, which acts as both the key and value. 
Unlike the standard query-based transformer, our proposed design utilizes the image feature as the query with our global context embedding as the key and value. This unique design allows our model to predict two distinct layouts, a departure from prior methods that predict only a single layout.
%

%% file: figs/network.tex
\begin{figure*}[tp]
    \centering
    \includegraphics[width=\textwidth]{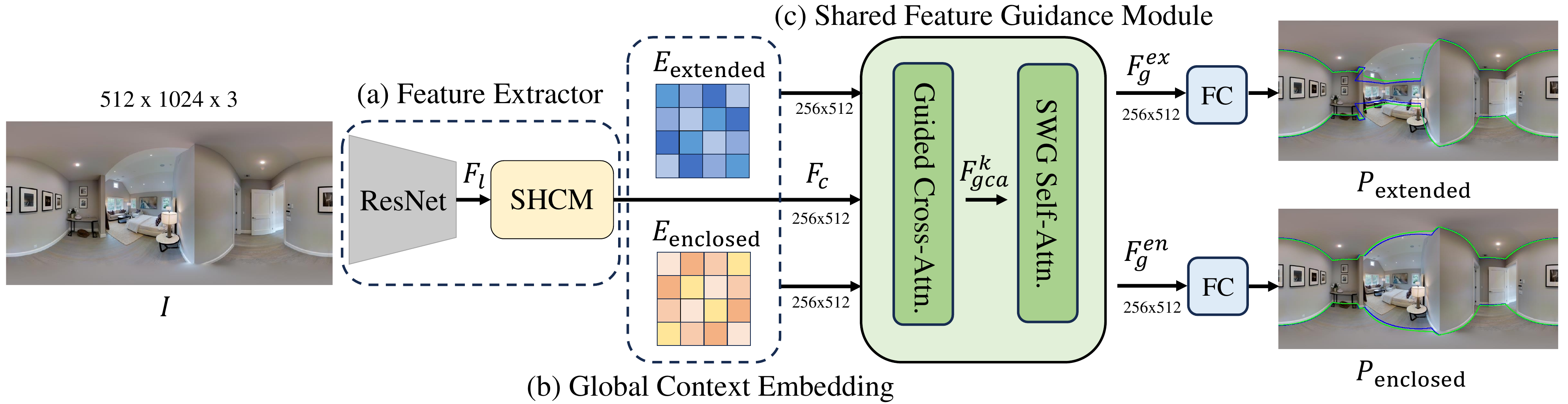}
    \vspace{-6mm}
    \caption{\textbf{Our Bi{-}Layout network architecture.} (a) \textbf{Feature extractor}: It processes a panoramic image $I$ using ResNet-50 to extract multi-scale features $F_{l}$ and then feeds those features into the Simplified Height Compression Module (SHCM) to produce the final compressed feature $F_{c}$. (b) \textbf{Global Context Embedding}: It consists of two learnable embeddings $E_{k}$, each designed to capture and encode the contextual information inherent in the corresponding type of layout labels. (c) \textbf{Shared Feature Guidance Module}: It consists of two components: Guided Cross-Attention and SWG Self-Attention. It guides the fusion of compressed feature $F_{c}$ with the global context embedding $E_{k}$ to generate feature $F_{g}^{k}$ ($k \in [\textit{extended},~\textit{enclosed}]$) more aligned for the corresponding layout type. Finally, we use fully connected (FC) layers to map $F_{g}^{k}$ to horizon-depth and room height, which are further converted to boundary layouts ($P_\text{extended}$ and $P_\text{enclosed}$).
    }
    \vspace{-5mm}
    \label{fig:network}
\end{figure*}

%% file: 04_method.tex
\section{Inherent Ambiguity in Labeled Data}
\label{sec:define_ambiguity}
We systematically examine instances of low IoU in the SoTA methods~\cite{wang2021led2,jiang2022lgt,shen2023dopnet} using MatterportLayout dataset~\cite{zou2021manhattan}. Our analysis identifies two types of ambiguity. 
First, when a \textit{enclosed} type GT label is given, the SoTA methods predict regions located outside of that designated room (See Fig.~\ref{fig:motivation}(a)). Conversely, when a \textit{extended} type GT label is given, the SoTA methods concentrate on the room where the camera is positioned (See Fig.~\ref{fig:motivation}(b)). These findings underscore inherent ambiguity within the testing GT labels. Moreover, since the SoTA models will predict either \textit{enclosed} or \textit{extended} types of layouts, the same ambiguity is likely to be within the training GT labels as well. This presents a substantial challenge for single-prediction-based SoTA methods.

\section{Method}
\label{sec:proposed_method}
To address the ambiguity issue in the dataset labels, we introduce our Bi{-}Layout model as shown in Fig.~\ref{fig:network}, which can generate two types of layout predictions $P_\text{extended}$ and $P_\text{enclosed}$. 
Our model mainly consists of three modules: feature extractor (Sec.~\ref{sec:extractor}), global context embedding (Sec.~\ref{sec:embedding}), and shared feature guidance module (Sec.~\ref{sec:guidance}). We describe each component and the training objectives (Sec.~\ref{sec:training_objective}) used for training our Bi{-}Layout model in the following sections.
%




\subsection{Feature Extractor}
\label{sec:extractor}
The feature extractor in our Bi{-}Layout model is shown in Fig.~\ref{fig:network}(a). We follow previous works~\cite{sun2019horizonnet,wang2021led2,jiang2022lgt,shen2023dopnet} to use ResNet-50~\cite{he2016deep} architecture to extract 2D image features $F_{l}$ of 4 different scales from the input panorama $I$. For each feature scale, we modify the module from~\cite{sun2019horizonnet} as \textit{Simplified Height Compression Module} (SHCM) to compress the features along the image height direction and generate 1D feature of the same dimension $\mathbb{R}^{N \times \frac{D}{4}}$, where $N$ is the width of feature map and $D$ is the feature dimension. Finally, we concatenate these features from different scales to generate the final compressed feature $F_{c}\in \mathbb{R}^{N \times {D}}$, where $N=256$ and $D=512$.
%

In contrast to previous works~\cite{sun2019horizonnet,wang2021led2,jiang2022lgt} setting $D=1024$, our design reduces model parameters.
By pruning our feature representation, we enhance the model's efficiency without compromising its effectiveness. To assess this design choice, we present detailed ablation studies in Sec.~\ref{sec:exp_ablation}.
%

\subsection{Global Context Embedding}
\label{sec:embedding}
Once the compressed feature $F_{c}$ is extracted, we introduce a novel and learnable embedding mechanism termed as \textit{Global Context Embedding}. This mechanism captures and encodes the overarching contextual information in a specific layout label, as illustrated in Fig.~\ref{fig:network}(b). We employ two learnable embeddings $E_{k} \in \mathbb{R}^{N \times {D}}$ where $k \in [\textit{extended},~\textit{enclosed}]$, one for \textit{extended} and the other one for \textit{enclosed} type. During training, these embeddings learn and encode diverse contextual information associated with different types of layout annotation. Moreover, they play a vital role in providing the dataset's global context information when queried by the compressed image feature $F_{c}$ via cross-attention in our shared feature guidance module (refer to Sec.~\ref{sec:guidance}). By infusing our compressed feature $F_{c}$ with this rich layout type-related embedding $E_{k}$, we generate diverse and meaningful predictions ($P_\text{extended}$ and $P_\text{enclosed}$), each aligned with a distinct global context of the dataset label.
%


\subsection{Shared Feature Guidance Module}
\label{sec:guidance}
\input{figs/guidance}
Building upon the compressed image feature $F_{c}$ and the global context embeddings $E_{k}$, as shown in Fig.~\ref{fig:network}(c), we present an innovative component called \textit{Shared Feature Guidance Module} (SFGM). This module can effectively guide the fusion of the image feature with the target global context embedding. Specifically,
we share the compressed image feature $F_{c}$ and use different global context embeddings $E_{k}$ (one at a time) as the inputs for our shared feature guidance module $SFGM(\cdot)$ to generate corresponding guided feature $F_{g}^{k} \in \mathbb{R}^{N \times {D}}$, denoted as: 
%
\begin{equation}
F_{g}^{k} = SFGM(F_{c}, E_{k}), ~k \in [\textit{extended},~\textit{enclosed}].
\label{eq:guidance}
\end{equation}
%
Our shared feature guidance module consists of \textit{Guided Cross-Attention} and \textit{SWG Self-Attention} as the building blocks, and the details of the architecture are shown in Fig.~\ref{fig:guidance}.

The standard cross-attention setting in DETR~\cite{carion2020end} or other high-level tasks~\cite{cheng2021per,cheng2022masked,meinhardt2022trackformer,zhou2022global,yue2023connecting,su2023slibo} uses embeddings as the query Q to retrieve relevant information from the corresponding image feature, which acts as both the key $\mathbf{K}$ and value $\mathbf{V}$ in order to generate the target outputs.
%
In our scenario, if we adopt the standard $\mathbf{QKV}$ setting, the shared image feature $F_{c}$ alone may not carry sufficient information to distinguish between the two types of distinct layouts. Therefore, we reverse this relationship and use our global context embeddings $E_{k}$ to learn from corresponding labels, guiding the image feature $F_{c}$ to generate the desired layout types.
%
%

As shown in Fig.~\ref{fig:guidance}, we use the compressed feature $F_{c}$ as the query $\mathbf{Q}$ and our global context embedding $E_{k}$ as both the key $\mathbf{K}$ and value $\mathbf{V}$ in our guided cross-attention $GCA(\cdot)$:
%
\begin{equation}
\resizebox{0.91\hsize}{!}{%
$
\begin{aligned}
\mathbf{Q} &= (F_{c}+PE_\text{sin})\mathbf{W}_{q}, \quad
\mathbf{K} = (E_{k}+PE_\text{learn})\mathbf{W}_{k},
\\
\mathbf{V} &= (E_{k}+PE_\text{learn})\mathbf{W}_{v}, \quad
F_{gca}^{k} = GCA(\mathbf{Q},\mathbf{K},\mathbf{V}),~
\label{eq:qkv}
\end{aligned}
$}
\end{equation}
where $F_{gca}^{k} \in \mathbb{R}^{N \times {D}}$ is the output of our guided cross-attention block. We apply different positional encoding strategies for these features, utilizing learnable positional encoding~\cite{carion2020end} $PE_\text{learn}\in \mathbb{R}^{N \times {D}}$ for our global context embedding $E_{k}$ and sinusoidal positional encoding~\cite{parmar2018image,bello2019attention} $PE_\text{sin}\in \mathbb{R}^{N \times {D}}$ for the compressed image feature $F_{c}$.
Each feature is then multiplied by its respective learnable weights $\mathbf{W}_{q/k/v}\in \mathbb{R}^{D\times D}$.
This unique design choice enables us to enrich the image feature $F_{c}$ by effectively incorporating our global context embedding $E_{k}$. 
Subsequently, this enriched feature $F_{gca}^{k}$ is served as $\mathbf{QKV}$ inputs to the SWG self-attention module~\cite{jiang2022lgt} for further enhancement.
As demonstrated in~\cite{jiang2022lgt}, the SWG self-attention module can effectively establish local and global geometric relationships within the room layout. Then, the process of guided cross-attention and SWG self-attention is repeated several times to refine the image feature with our context embeddings to generate the final guided feature $F_{g}^{k}$, as shown in Fig.~\ref{fig:guidance}.

By employing this novel cross-attention design, we achieve an enriched and context-aware guided feature representation $F_{g}^{k}$ that is subsequently utilized for generating our Bi{-}Layout predictions ($P_\text{extended}$ and $P_\text{enclosed}$), each possessing distinct and valuable properties. This flexibility in our method enables us to provide diverse layout predictions tailored to different global context embeddings $E_{k}$ and input panorama features $F_{c}$.
Built on this architectural design, our model can be compact and efficient to generalize to more label types by increasing global context embeddings $E_{k}$.
%

\subsection{Training Objective}
\label{sec:training_objective}
After obtaining the target feature $F_{g}^{k}$, we follow~\cite{jiang2022lgt,shen2023dopnet} using fully connected (FC) layers to map the feature $F_{g}^{k}$ to horizon-depth $d_{k}=\{d_k^i\}_{i=1}^{N}$ and room height $h_{k}$, where $N$ is the width of the feature map. We can apply the explicit transformation to convert the horizon depth and room height to layout boundaries $P_{k}$ on the panorama. We further convert column-wise depth $d_{k}^i$ into depth normal $n_{k}^i$ and gradient of normal $g_{k}^{i}$, $k \in [\textit{extended},~\textit{enclosed}]$.

The loss functions for depth, normal, gradient, and room height are defined as follows:
\vspace{-1mm}
\begin{equation}
\resizebox{0.91\hsize}{!}{%
$
\begin{aligned}
\mathcal{L}_\text{depth}^{k} &= \frac{1}{N} \sum_{i \in N} |d_k^{i}-d_\text{gt}^{i}|,~
\mathcal{L}_\text{normal}^{k} = \frac{1}{N} \sum_{i \in N}(- n_k^{i} \cdot {n}_\text{gt}^{i}),
\\
\mathcal{L}_\text{gradient}^{k} &= \frac{1}{N} \sum_{i \in N} |g_k^{i}-{g}_\text{gt}^{i}|,~
\mathcal{L}_\text{height}^{k} = |h_k-h_\text{gt}|,
\label{eq:branch_loss}
\end{aligned}
$}
\vspace{-1mm}
\end{equation}
where $d_\text{gt}^{i}, {n}_\text{gt}^{i}, {g}_\text{gt}^{i}$, and $h_\text{gt}$ denote the ground truth of depth, normal, gradient, and room height respectively. We calculate the L1 loss for depth loss, gradient loss, height loss, and cosine similarity for normal loss.
Our branch loss $\mathcal{L}_{k}$ is:
%
\begin{equation}
\resizebox{0.91\hsize}{!}{%
$
\mathcal{L}_{k} = \lambda_d\mathcal{L}_\text{depth}^{k}+\lambda_n\mathcal{L}_\text{normal}^{k}+\lambda_g\mathcal{L}_\text{gradient}^{k}+\lambda_h\mathcal{L}_\text{height}^{k},
\label{eq:total_loss}
$}
\end{equation}
where $k \in [\textit{extended},~\textit{enclosed}]$ and the final loss $\mathcal{L}_\text{total}=\mathcal{L}_\text{extended}+\mathcal{L}_\text{enclosed}$. We set $\lambda_d = 0.9$, $\lambda_n = 0.1$, $\lambda_g = 0.1$ and $\lambda_h = 0.1$ for both branches to balance the model weight.

%% file: figs/guidance.tex
\begin{figure}[tp]
    \centering
    \includegraphics[width=\linewidth]{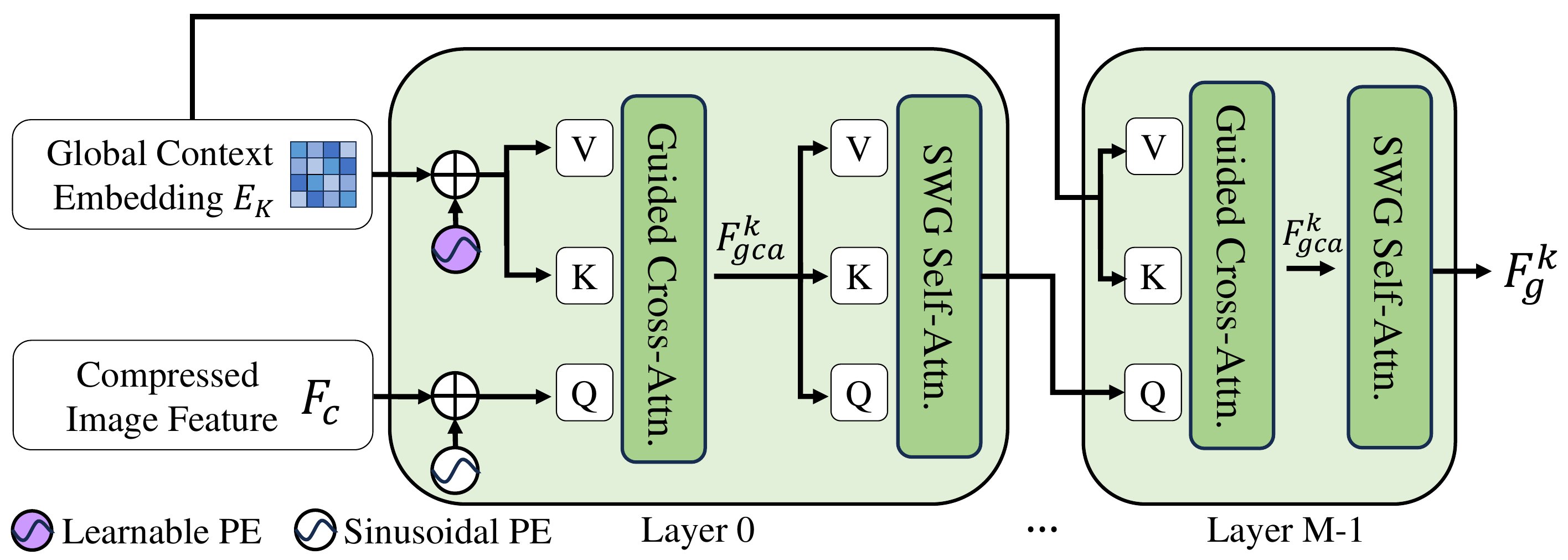}
    \vspace{-4mm}
    \caption{\textbf{Our Shared Feature Guidance Module architecture (SFGM).} It consists of two blocks: Guided Cross-Attention and SWG Self-Attention. The module has $M=8$ layers, and the structure of each layer is identical.
    Given the compressed image feature $F_{c}$ and global context embedding $E_{k}$, we first apply the sinusoidal and learnable positional encoding, respectively. With the compressed feature $F_{c}$ as the query $\mathbf{Q}$ and our global context embedding $E_{k}$ as both the key $\mathbf{K}$ and value $\mathbf{V}$, our guided cross-attention generates the feature $F_{gca}^{k}$, and it is served as $\mathbf{QKV}$ inputs to SWG self-attention. This process will repeat and further refine the output feature with our global context embedding to generate the final guided feature $F_{g}^{k}$.}
    \vspace{-4mm}
    \label{fig:guidance}
\end{figure}

%% file: 08_experiments.tex
\section{Experiments}
\label{sec:exp}
We conduct our experiments on a single NVIDIA RTX 4090 GPU and implement the proposed method with PyTorch~\cite{paszke2019pytorch}. For training, we use a batch size of 12 and set the learning rate to $1 \times 10^{-4}$. We select Adam as our optimizer, adhering to its default configurations. For the data augmentation, we use the technique proposed in~\cite{sun2019horizonnet}, including left-right flipping, panoramic horizontal rotation, luminance adjustment, and panoramic stretch.

\subsection{Datasets}
\label{sec:exp_datasets}

\paragraph{MatterportLayout.}
\label{sec:exp_mp3d}
MatterportLayout~\cite{zou2021manhattan} contains {2295} samples labeled by Zou~\etal~\cite{zou2021manhattan}.
However, as we analyzed in Sec.~\ref{sec:define_ambiguity}, this dataset has annotation ambiguity, and many images with ambiguity are annotated with the \textit{extended} type. 
Hence, we propose a semi-automatic procedure (provide the details in supplementary material) to re-annotate \textit{enclosed} type labels from the ambiguous \textit{extended} ones. We re-annotate $15\%$ of the labels in the whole dataset.
Note that these new labels of \textit{enclosed} type plus the remaining $85\%$ of original labels will be used to train our Bi-Layout model's \textit{enclosed} branch. In contrast, all original labels will be used to train the \textit{extended} branch.
For a fair comparison with SoTA methods, we use the original label and the same testing split for evaluation.

\vspace{-2mm}
\paragraph{ZInD.}
\label{sec:exp_zind}
ZInD~\cite{cruz2021zillow} dataset is currently the largest dataset with room layout annotations. It provides both \textit{raw} and \textit{visible} labels, which is similar to our defined \textit{enclosed} and \textit{extended} types, respectively.
Besides, ZInD separates the data into \textit{simple} and \textit{complex} subsets based on whether the images have contiguous occluded corners. 
Therefore, we have two variants of ZInD in our experiments: 
\textbf{(a) ZInD-Simple} represents the \textit{simple} subset and consists of {24882}, {3080}, and {3170} panoramas for training, validation, and testing splits.
\textbf{(b) ZInD-All} represents the whole dataset with {50916}, {6352}, and {6352} panoramas for each split. It has complex opening regions, resulting in more severe ambiguity issues.
Therefore, it can better evaluate the robustness of different methods for handling the ambiguity issue.
In both ZInD dataset variants, we use the \textit{raw} and \textit{visible} labels to train our \textit{enclosed} and \textit{extended} branches, respectively, and follow the SoTA methods to test on \textit{raw} labels for a fair comparison.

\subsection{Disambiguate Metric}
\label{sec:disambiguate} 
If we already know the layout type of each test image, we can use this information to select the output from the corresponding branch for evaluation. However, we find that the test data has annotation ambiguity; even the \textit{raw} labels in ZInD are still not exempt from this issue. 

To address the above issue and demonstrate our model's capability to handle the annotation ambiguity, we introduce a new metric, termed the \textit{disambiguate} metric, as follows:
\vspace{-2mm}
\begin{equation}
IoU_{\text{disambiguate}} = \sum_{i=0}^{S} \arg \max_{P_{k}^{i}\in \mathbb{P}} IoU(P_{\text{gt}}^{i},~P_{k}^{i}),
\label{eq:tolerance}
\end{equation}
where $P_{k}^{i}\in \mathbb{P},~k\in [\textit{extended},~\textit{enclosed}]$ denotes layout predictions from both branches and $P_{gt}^{i}$ denotes the ground truth layout. We first calculate the Intersection over Union (IoU) between each prediction and ground truth (GT) for each image and then select the higher IoU for averaging all samples. This is because the higher IoU serves as the disambiguate prediction and represents the most suitable prediction when encountering ambiguity. 

The proposed metric effectively provides a robust and quantitative measure of how a method excels in handling ambiguous scenarios within the dataset without necessitating manual corrections to the ambiguous annotations. In other words, we can use the labels provided by the original dataset to do the evaluation.
\input{tables/all_table}

\subsection{Comparison with State-of-the-Art Methods}
\label{sec:exp_comparison_SOTA}
\vspace{-2mm}
\paragraph{Evaluation settings.}
Since our model outputs two layouts, we propose to compare our method with the SoTA methods in two ways.
\textbf{Using the equivalent branch}: We use the output from the branch trained with the same data as other methods. Specifically, we use the output from \textit{extended} branch for the MatterportLayout dataset and the output from \textit{enclosed} branch for ZInD to fairly compare with other methods.
\textbf{Using both branches with disambiguate metric}: We use the proposed \textit{disambiguate} metric as defined in Sec.~\ref{sec:disambiguate} to evaluate the performance of our method.
\vspace{-4mm}
\paragraph{Full set evaluation.}
We present quantitative results on three different datasets in Table~\ref{tab:all_table}(a). 
As the SoTA methods do not experiment on the ZInD-All dataset, we retrain all baseline models based on their official repositories. 
The results demonstrate that our \textit{equivalent} branch consistently outperforms SoTA methods across all datasets, underscoring the advantages of joint training with bi-layout data.
Furthermore, our disambiguated results surpass these benchmarks, affirming the existence of ambiguity in the original annotations.
Our Bi{-}Layout model effectively mitigates this issue by selecting the most appropriate prediction. Notably, in terms of model size, our architecture, despite generating bi-layout predictions, maintains a smaller total parameter size compared to the SoTA models. This underscores the efficiency of our design in achieving superior performance with a more compact model.

\vspace{-4mm}
\paragraph{Subset evaluation.}
To highlight the ambiguity issue, we select a subset based on the failure predictions of all the previous SoTA models~\cite{wang2021led2,jiang2022lgt,shen2023dopnet}.
For each SoTA model, we find images with the 2DIoU evaluation lower than $0.6$. 
Next, we combine all these failure cases among all the SoTA models to construct the subset. 
Finally, the subsets consist of $11\%$, $6\%$, and $18\%$ of the test data in three datasets, respectively.
The quantitative results in Table~\ref{tab:all_table}(b) reveal a more pronounced performance gap between our method and the SoTA models, with differences reaching up to $9.28\%$ in 2DIoU on the most ambiguous ZInD-All dataset.
We also provide the qualitative results in Fig.~\ref{fig:qualitative_comparison},
where a bird's-eye view of the predictions vividly illustrates the significant challenges posed by ambiguity.
This confirms that layout ambiguity is a key cause for low IoUs of previous methods, and our Bi-Layout estimation is effective in addressing the issue and performs remarkably well in this subset.

\input{figs/subset}

\subsection{Ambiguity Detection}
\input{figs/opening}

Our Bi-Layout model can naturally detect ambiguous regions by comparing the per-column pixel difference between two predicted layout boundaries.
This per-column pixel difference can serve as our predicted confidence score.
If the difference is larger, the column is more likely to be an ambiguous region (i.e., typically an opening room structure).
We formulate the detection of ambiguous regions into a binary classification task where GT ambiguous regions are columns with more than 2 pixels' difference between two annotations of the \textit{extended} and \textit{enclosed} types. We predict columns with more than 10 pixels' difference between predicted layouts as ambiguous regions.
We test this method on ZInD as it is the only dataset that provides both types of GT labels (\textit{i.e., raw} and \textit{visible}) in testing. 
Our method achieves a reasonably high Precision of $\textbf{0.82}$ and Recall of $\textbf{0.71}$.
Moreover, the qualitative results in Fig.~\ref{fig:opening} further demonstrate that our method can indeed detect ambiguous regions. We believe this is particularly useful for applications where the model can highlight ambiguous regions and let the users select suitable predictions for their use cases.

\subsection{Abalation Studies}
\label{sec:exp_ablation}
\vspace{-2mm}

\paragraph{Different feature fusion designs.}
To fuse the information from image features and our global context embeddings, we conduct comprehensive ablation studies to validate the effectiveness of different designs. 
The fusion methods include add, concatenation, AdaIn~\cite{huang2017arbitrary}, and FiLM~\cite{perez2018film}.
We further investigate two Query-Key-Value (QKV) feature designs in our shared feature guidance module. 

The results in Table~\ref{tab:qkv_order} show that our proposed design significantly outperforms all feature fusion methods and the standard cross-attention setting, where the global context embedding $E_{k}$ served as query $\mathbf{Q}$. The compressed image feature $F_{c}$ served as key $\mathbf{K}$ and value $\mathbf{V}$. This demonstrates that our design can effectively utilize contextual information within the embedding $E_{k}$ to enhance the alignment of the compressed feature $F_{c}$ with the corresponding layout prediction type.

\input{figs/architecture_comparison}
\input{tables/qkv_order}
\vspace{-4mm}
\paragraph{Comparions of model architectures.}
As shown in Fig.~\ref{fig:architecture_comparison}, we compare three model architectures to predict multiple layouts. We conduct the quantitative comparison in Table~\ref{tab:parameter}. 
\noindent \textbf{Two models}: Training two models of the same architecture to predict two layout types is the most straightforward design (Fig.~\ref{fig:architecture_comparison}(a)). This architecture achieves good performance without interference between learning two types of layout. However, this doubles the model size and training time. 

\noindent \textbf{Two transformers}: An alternative is to share the feature extractor but separate the transformer and prediction head (Fig.~\ref{fig:architecture_comparison}(b)).
This saves model size a little, but the performance drops significantly as it cannot handle the interference when learning two types of layouts simultaneously.

\noindent \textbf{Our model}: Our model is the smallest as we share both the feature extractor and transformer and only separate the lightweight prediction head (Fig.~\ref{fig:architecture_comparison}(c)). To reduce the interference in learning two layouts, we introduce two learnable global context embeddings, which can inject layout type-related context information into the image feature via cross-attention. Therefore, our model achieves comparable or better performance than others. In addition, there is only a slight performance drop if we further reduce the model size by reducing the compressed feature channel dimension from {1024} to {512}. Our final model ($c=512$) balances performance and parameter efficiency best.

\input{tables/parameter}


%% file: tables/all_table.tex
\begin{table*}[t]
\vspace{-4mm}
\centering
  \resizebox{\linewidth}{!} 
  {
    \centering
    \begin{tabular}{lccccccc}
    \toprule
        \textbf{(a) Full set}& & \multicolumn{2}{c}{MatterportLayout~\cite{zou2021manhattan}} & \multicolumn{2}{c}{ZInD-Simple~\cite{cruz2021zillow}} & \multicolumn{2}{c}{ZInD-All~\cite{cruz2021zillow}} \\
    \midrule
    Method & \# Params & 2DIoU(\%) & 3DIoU(\%) & 2DIoU(\%) & 3DIoU(\%) & 2DIoU(\%) & 3DIoU(\%) \\
    \midrule
    LED${^2}$Net~\cite{wang2021led2} & {82 M} & 82.37 & 80.05 & 90.20 & 88.34 & 82.31 & 80.28 \\
    LGT-Net~\cite{jiang2022lgt} & 136 M & 83.52 & 81.11 & 91.77 & 89.95 & 84.07 & 82.09 \\
    DOP-Net~\cite{shen2023dopnet} & 137 M & 84.11 & 81.70  & 91.94 & 90.13 & 83.92 & 81.87 \\
    \midrule
    \rowcolor{ourcolor} 
    Ours (equivalent branch) & 102 M & 84.56 & 82.05 & 92.07 & 90.25 & 84.90 & 82.96 \\
    \rowcolor{ourcolor} 
    Ours (disambiguate) & 102 M & \textbf{85.10} & \textbf{82.57} & \textbf{92.79} & \textbf{90.95} & \textbf{86.21} & \textbf{84.22} \\
    \bottomrule
    \end{tabular}%
  }
  \centering
  \resizebox{\linewidth}{!} 
  {
  \centering
    \begin{tabular}{lccccccc}
    \toprule
          \textbf{(b) Subset}& & \multicolumn{2}{c}{MatterportLayout~\cite{zou2021manhattan}} & \multicolumn{2}{c}{ZInD-Simple~\cite{cruz2021zillow}} & \multicolumn{2}{c}{ZInD-All~\cite{cruz2021zillow}} \\
    \midrule
    Method & \# Params & 2DIoU(\%) & 3DIoU(\%) & 2DIoU(\%) & 3DIoU(\%) & 2DIoU(\%) & 3DIoU(\%) \\
    \midrule
    LED${^2}$Net~\cite{wang2021led2} & 82 M & 53.57  & 51.12  & 45.31 & 44.10  & 48.76 & 47.35 \\
    LGT-Net~\cite{jiang2022lgt} & 136 M & 53.17 & 50.54 & 53.20 & 52.00 &  50.89 & 49.58 \\
    DOP-Net~\cite{shen2023dopnet} & 137 M & 57.13 & 54.80 & 51.55 & 50.26 &  50.92  &  49.46 \\
    \midrule
    \rowcolor{ourcolor} 
    Ours (equivalent branch) & 102 M & 59.85 & 57.08 & 55.09 & 53.76 & 54.22  & 52.78 \\
    \rowcolor{ourcolor} 
    Ours (disambiguate) & 102 M & \textbf{62.81} & \textbf{59.97} & \textbf{62.10} & \textbf{60.63} & \textbf{60.20}  & \textbf{58.53} \\
    \bottomrule
    \end{tabular}%
  }
  \vspace{-2mm}
  \caption{\textbf{Full set and Subset evaluation.} 
  \textbf{Equivalent branch} represents the output, which is trained with the same label as baseline methods. \textbf{Disambiguate} is our proposed metric.}
  \vspace{-4mm}
  \label{tab:all_table}%
\end{table*}%

%% file: figs/subset.tex
\begin{figure*}
  \centering
  \vspace{-4mm}
  \begin{subfigure}{1\linewidth}
    \includegraphics[width=1\linewidth]{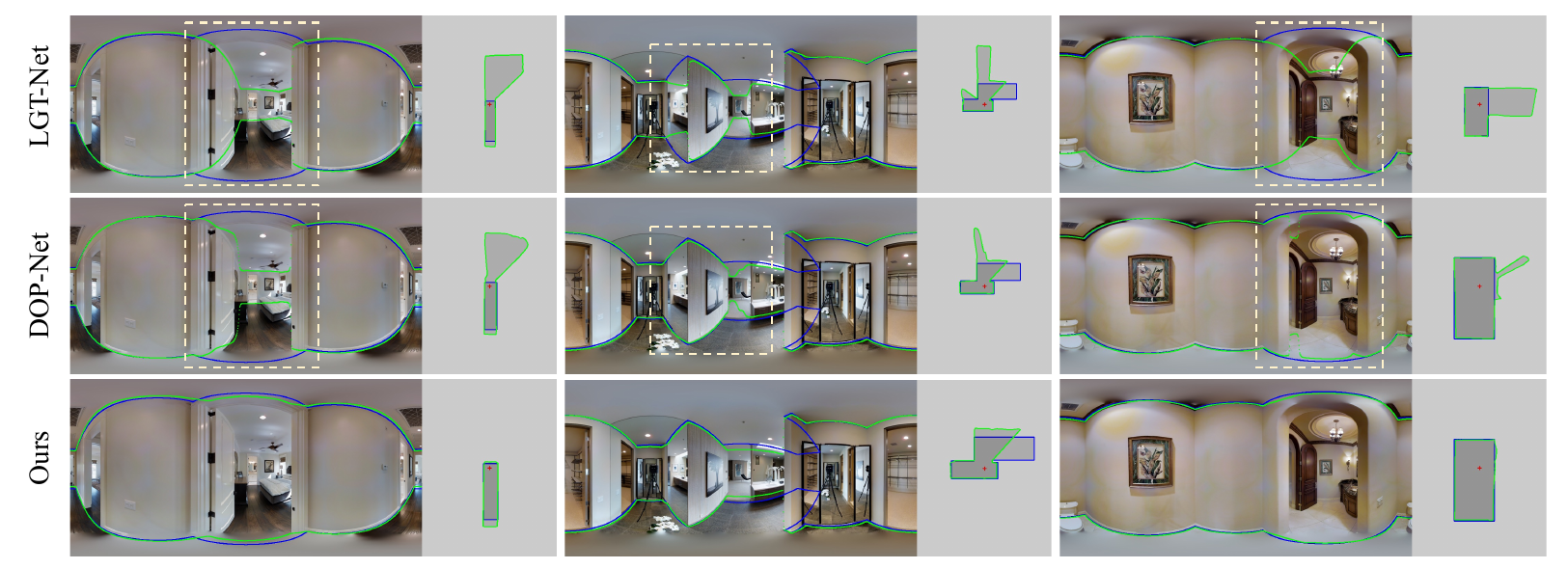}
    \caption{Qualitative comparison on the MatterportLayout~\cite{zou2021manhattan} dataset.}
    \label{fig:qualitative_comparison_mp3d}
  \end{subfigure}
  \\
  \begin{subfigure}{1\linewidth}
    \includegraphics[width=1\linewidth]{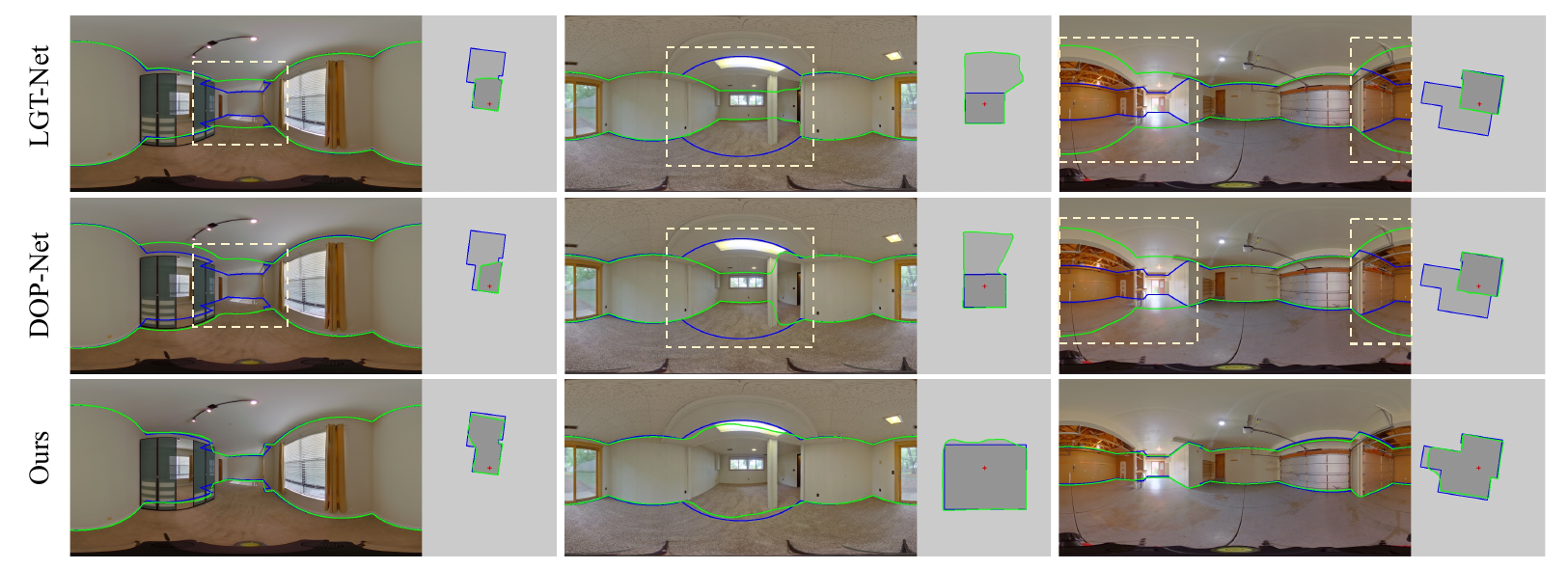}
    \setlength{\belowcaptionskip}{-2mm}\centering\caption{Qualitative comparison on the ZInd~\cite{cruz2021zillow} dataset.}
    \label{fig:qualitative_comparison_zind}
  \end{subfigure}
  \caption{
  \textbf{Qualitative comparison} on the MatterportLayout~\cite{zou2021manhattan} (top) and  ZInd~\cite{cruz2021zillow} datasets (bottom). \blue{Blue} and \green{Green} represent ground truth labels and predictions, respectively. The boundaries of the room layout are on the left, and their bird's eye view projections are on the right. 
  We show our \textit{disambiguate} results, which effectively address the ambiguity issue, while the SoTA methods struggle with the ambiguity, as highlighted in dashed lines.}
  \label{fig:qualitative_comparison}
   \vspace{-4mm}
\end{figure*}

%% file: figs/opening.tex
\begin{figure}[tp]
  \centering
  \includegraphics[width=0.95\linewidth]{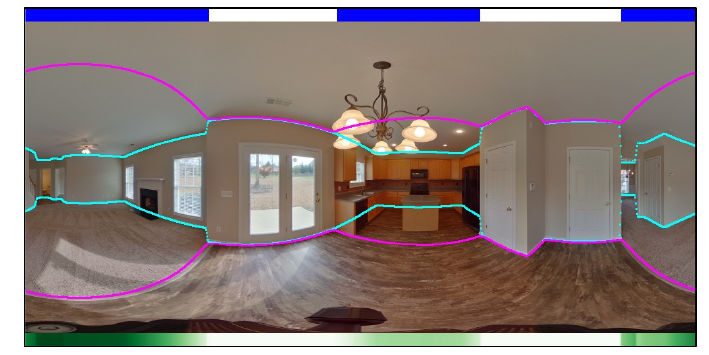}
  \vspace{-2mm}
  \caption{\textbf{Qualitative results for ambiguity detection.} \blue{Blue} and \green{Green} on the top and bottom rows represent ground truth and predicted confidence, respectively. \textcolor{cyan}{Cyan} and \textcolor{magenta}{Megenta} lines are our \textit{extended} and \textit{enclosed} type layout predictions. 
  With these two predictions, our model can accurately detect ambiguous regions.
  } 
  \vspace{-6mm}
  \label{fig:opening}
\end{figure}

%% file: figs/architecture_comparison.tex
\begin{figure}[tp]
    \centering
    \includegraphics[width=\linewidth]{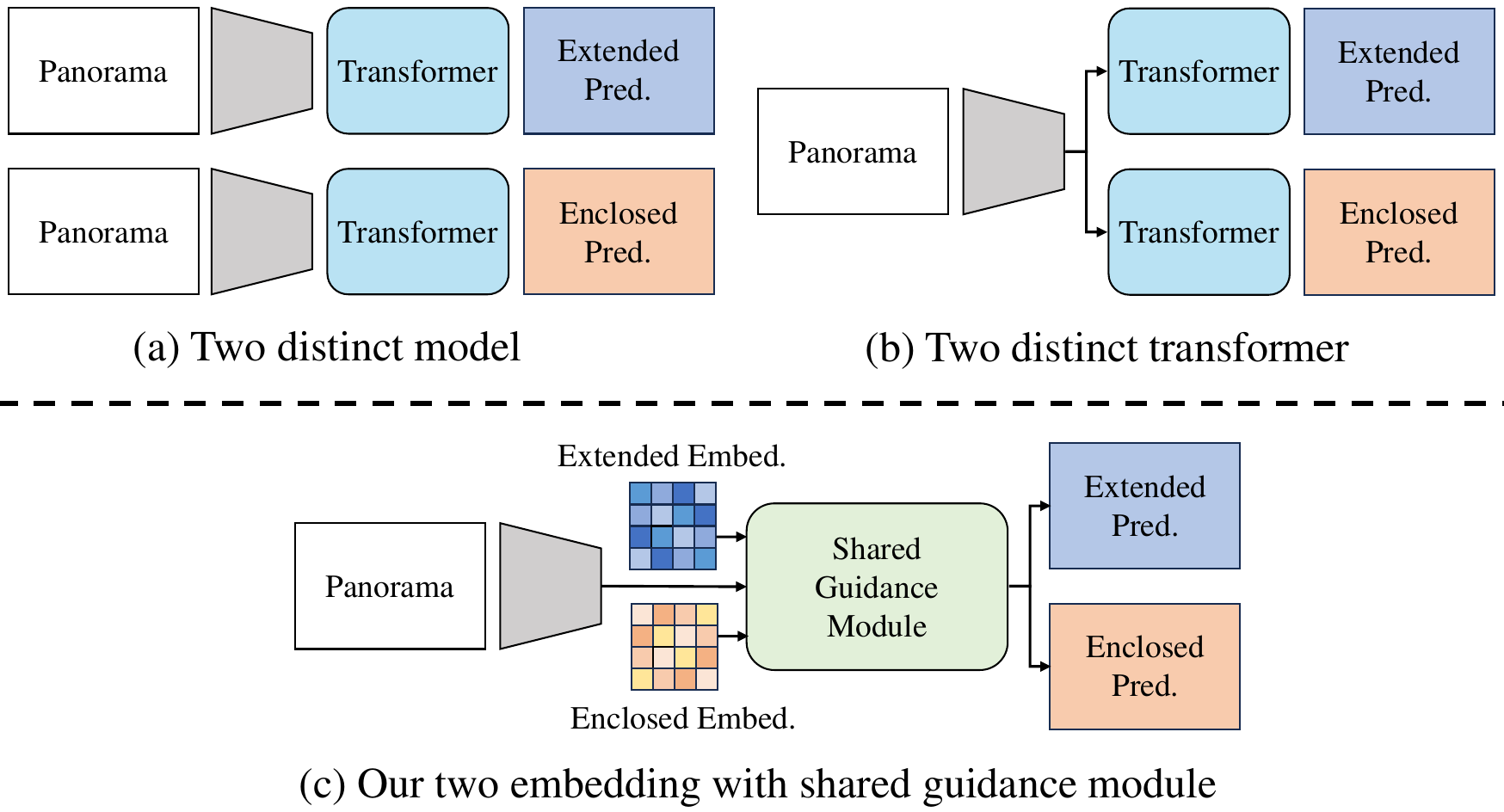}
    \vspace{-4mm}
    \caption{\textbf{Model architecture comparison.} We show the different model architecture designs for predicting multiple layouts. Please refer to Table~\ref{tab:parameter} for quantitative comparison. }
    \vspace{-4mm}
    \label{fig:architecture_comparison}
\end{figure}

%% file: tables/qkv_order.tex
\begin{table}[tp]
  \centering
    \begin{tabular}{lcccc}
    \toprule
    Design &Q & KV & 2D IoU (\%) & 3D IoU (\%) \\
    \midrule
    Add &  &  & 83.79  & 81.26 \\
    Concat &  &  &  84.17  &  81.44 \\
    AdaIn~\cite{huang2017arbitrary} &  &  &  83.28  &  80.54 \\
    FiLM~\cite{perez2018film} &  &  &  84.08  & 81.39  \\
    \midrule
    Standard&$E_{k}$& $F_{c}$ & 83.34  & 80.85 \\
    \rowcolor{ourcolor} 
    Ours &$F_{c}$& $E_{k}$&\textbf{85.10}&\textbf{82.57}\\
    \bottomrule
    \end{tabular}%
  \vspace{-2mm}
  \caption{\textbf{Comparison of QKV feature designs.} 
  $F_{c}$ is the compressed feature, and $E_{k}$ is our global context embedding.
  We evaluate different designs of QKV features in the Shared Feature Guidance Module on MatterportLayout~\cite{zou2021manhattan} with our proposed \textit{disambiguate} metric. }
  \vspace{-2mm}
  \label{tab:qkv_order}%
\end{table}%

%% file: tables/parameter.tex

\begin{table}[tp]
\centering
  \resizebox{\linewidth}{!} 
  {
  \centering
    \begin{tabular}{lccc}
    \toprule
    Method & \# Params & 2DIoU(\%) & 3DIoU(\%) \\
    \midrule
    Two models & 272 M  & \textbf{85.29} & 82.72 \\
    Two Transformers & 203 M  & 84.35 & 81.88 \\
    \midrule
    \rowcolor{ourcolor} 
    Ours (c = 1024) & 172 M  & 85.25 & \textbf{82.76}\\
    \rowcolor{ourcolor}
    Ours (c = 512) &   \textbf{102 M}    &  85.10    &  82.57\\
    \bottomrule
    \end{tabular}%
  }
  \vspace{-2mm}
  \caption{\textbf{Model size and performance trade-off.} In this part, we only compare to the LGT-Net~\cite{jiang2022lgt} model variances (i.e., The first two rows) since our model is built on top of its architecture. In the third row, c represents the \# channel of the image feature. We conduct these quantitative evaluations on MatterportLayout~\cite{zou2021manhattan} with our proposed \textit{disambiguate} metric. Our final model strikes the best balance between performance and parameter efficiency.}
  \vspace{-6mm}
  \label{tab:parameter}%
\end{table}%

%% file: 10_conclusion.tex
\section{Conclusion}
\label{sec:conclusion}

We propose a novel Bi{-}Layout model to generate two distinct predictions, effectively resolving the layout ambiguity.
Most importantly, we introduce a novel embedding mechanism with a shared feature guidance module, where each embedding is designed to learn the global context inherent in each type of layout. Our model strikes a good balance between model compactness and prediction accuracy with these designs. In addition, we propose a \textit{disambiguate} metric to evaluate the accuracy with multiple predictions. On MatterportLayout~\cite{zou2021manhattan} and ZInD~\cite{cruz2021zillow} datasets, our method outperforms other state-of-the-art methods, especially on the subset setting with considerable ambiguity.

%% file: 12_appendix.tex


\maketitlesupplementary

\section{Overview}
This supplementary material presents additional results to complement the main manuscript.
We first introduce our relabeling pipeline in Sec.~\ref{supp_relabel_pipeline} and show some bi{-}layout annotation examples in Sec.~\ref{supp_annotation_vis}. In Sec.~\ref{supp_compare_sota}, we provide more qualitative comparisons with the state-of-the-art (SoTA) methods. In Sec.~\ref{supp_detection}, we show more examples of our ambiguity detection under different scenarios to validate the robustness of our method. We conduct additional ablation studies in Sec.~\ref{supp_ablation} to compare our method in more comprehensive settings. Finally, we show the limitations in Sec.~\ref{supp_limitation} and provide some future research directions in Sec.~\ref{supp_future_work}.

\section{Semi-automatic Relabeling}
\label{supp_relabel_pipeline}
\input{figs/supp_relabel_flow}
We introduce our semi-automatic relabeling pipeline for annotating the second type of layout on the MatterportLayout~\cite{zou2021manhattan} dataset as follows and shown in Fig.~\ref{fig:supp_relabel_flow}:
\begin{enumerate}[label=(\alph*)]
\item Given the original annotations from the MatterportLayout~\cite{zou2021manhattan} dataset. We check each column of the panorama, if there are more than two annotations in the same column, we define it as the occlusion part. As shown in Fig.~\ref{fig:supp_relabel_flow}(a), \blue{blue line} indicates the original annotation, and the dashed lines highlight the occlusion region.
\item Next, we take the original annotation and project it to the bird’s-eye view floorplan coordinate, aligning it with the center of the camera. As shown in Fig.~\ref{fig:supp_relabel_flow}(b), the isolated \red{red point} indicates the center of the camera.
\item After obtaining the annotation on the floorplan coordinate, we categorize the corners as either \textit{visible} or \textit{invisible}, representing whether the corners can be seen from the center of the camera or not. We find the closest \textit{visible} points in the occlusion region as our candidate corners. As shown in Fig.~\ref{fig:supp_relabel_flow}(c), the \red{red boxes} indicate our candidate visible corners.
\item Once we have our candidate corners, we generate several annotation proposals using these points. As shown in Fig.~\ref{fig:supp_relabel_flow}(d), the \red{red lines} are our annotation proposals based on the candidate corners.
\item We select the best proposal, which should provide a clear boundary between different rooms. Note that this is the only step that needs a human decision. As shown in Fig.~\ref{fig:supp_relabel_flow}(e), we manually choose the proposal to separate the two rooms in this case.
\item Finally, we project these newly defined corners back to the panorama view, creating our relabeled annotation for the panorama. As shown in Fig.~\ref{fig:supp_relabel_flow}(f), \green{green line} indicates the relabeled annotation.
\end{enumerate}

We introduce our semi-automatic relabeling pipeline with the above steps, which clearly explain how we relabel the MatterportLayout~\cite{zou2021manhattan} dataset. With this relabel pipeline, we can generate the \textit{enclosed} type of annotations from the \textit{extended} type of annotations and use these new labels with the original labels to train our Bi{-}Layout model. 

\section{Bi-layout Annotations}
\label{supp_annotation_vis}
\input{figs/supp_relabel_results}
\input{figs/supp_zind_gt}

We show some bi-layout annotation examples in both the MatterportLayout~\cite{zou2021manhattan} and ZInD~\cite{cruz2021zillow} dataset.

\noindent\textbf{MatterportLayout} We present some cases of our relabeled annotations for the MatterportLayout~\cite{zou2021manhattan} dataset. The original annotations in Fig.~\ref{fig:supp_relabel_results}(a) are from the original dataset labels. Our relabeled annotations are shown in Fig.~\ref{fig:supp_relabel_results}(b). Based on our definition, we relabel the \textit{extended} type of annotation to the \textit{enclosed} type of annotation.

\noindent\textbf{ZInD}
We show some cases of two types of annotations officially provided by ZInD~\cite{cruz2021zillow} dataset. The \textit{visible} annotations are shown in Fig.~\ref{fig:supp_zind_gt}(a), and the \textit{raw} annotations are shown in Fig.~\ref{fig:supp_zind_gt}(b), which corresponds to our \textit{extended} type and \textit{enclosed} type, respectively.

\section{Comparisons with SoTA}
\label{supp_compare_sota}
We show more qualitative results on the MatterportLayout~\cite{zou2021manhattan} dataset in Fig.~\ref{fig:mp3d_supp} and the ZInD~\cite{cruz2021zillow} dataset in Fig.~\ref{fig:zind_supp}. Our Bi{-}Layout model can effectively address the ambiguity issue that the SoTA methods struggle with.

\input{figs/mp3d_supp}
\input{figs/zind_supp}

\section{Ambiguity Detection}
\label{supp_detection}
We show more qualitative results and several scenarios of ambiguity detection in Fig.~\ref{fig:supp_ambiguity_detection}. In (a) and (b), we provide more examples to demonstrate that our Bi{-}Layout model can accurately detect ambiguous regions as the GT shows. In (c), we offer a normal case where there is no ambiguous region in the image, and our model can predict two identical predictions. In (d), we show a special case where the GT does not indicate the ambiguous regions as it should be (i.e., GT itself has ambiguity), and our model can still successfully identify them, showing the capability of our method to address the inherent ambiguity issue in the dataset.
\input{figs/supp_ambiguity_detection}

\section{Ablation Studies}
\label{supp_ablation}
\noindent \textbf{Global Context Embedding.} 
To show the effectiveness of our proposed \textit{Global Context Embedding} and \textit{Shared Feature Guidance Module}, we conduct the experiment using a \textbf{\textit{single}} global context embedding for our feature guidance module and only generate a \textbf{\textit{single}} prediction as the conventional methods present (i.e., a single branch version of our proposed method). We compare our single branch with LGT-Net~\cite{jiang2022lgt} since the proposed components are built on top of its architecture.
In Table~\ref{tab:embedding_supp}, our single branch outperforms the baseline method on both the full set and subset of the MatterportLayout~\cite{zou2021manhattan} dataset, showing the effectiveness of our Global Context Embedding design.

\input{tables/supp_single_comparison}

\noindent \textbf{Model size comparison.} 
As discussed in the main manuscript, many model variations can let the SoTA method predict two layouts. We additionally show the \textit{two-head} version of the baseline model, which shares the feature extractor and transformer parts, and simply add the other prediction head to generate the second type of layout. We compare all the model variations in Table~\ref{tab:model_size_supp}. Although the \textit{two-head} version model decreases the model parameters, the performance is degraded significantly due to the naive model design.
The comparison with these model variations shows the effectiveness and compactness of our Bi{-}Layout model.

\input{tables/supp_model_variation}

\noindent \textbf{Image feature dimension.}
To make the model more compact, we compare different image feature dimensions: {1024}, {512}, and {256}. We experiment on the full set and subset of the MatterportLayout~\cite{zou2021manhattan} dataset. In Table~\ref{tab:our_diff_model_supp}, the feature dimension of {1024} performs the best but it has the largest model size. Our proposed Bi{-}Layout model with a feature dimension of {512} strikes a good balance between the performance and model size. When it comes to the feature dimension of {256}, the performance significantly drops, which means too small feature dimensions are not feasible for our task.

\input{tables/supp_parameter}


\paragraph{Pretrain with more data.}
MatterportLayout~\cite{zou2021manhattan} has limited bi-layout samples, with only $15\%$ of {1647} training images being re-annotated. However, most images in ZInD-Simple~\cite{cruz2021zillow} ({24,882}) and ZInD-All~\cite{cruz2021zillow} ({50,916}) have both \textit{raw} and \textit{visible} labels.
Although ZInD also has ambiguity issues, we believe pretraining on ZInD with extensive and diverse bi-layouts can boost the model's performance on MatterportLayout.
Therefore, we train one model from scratch and fine-tune two models pre-trained on ZInD-Simple and ZInD-All, respectively.
The results in Table~\ref{tab:pretraining} demonstrate that pretraining indeed aids the model in disambiguating, with a more significant performance gain observed when more data is used during the pretraining stage. This suggests that with additional bi-layout annotations, our model has the potential to more effectively address the ambiguity issue.
\input{tables/pretraining}




\section{Limitations}
\label{supp_limitation}
\input{figs/supp_failure_case}
Our Bi{-}Layout model also has limitations. As shown in Fig.~\ref{fig:supp_failure_case}, we provide the predictions from our two branches, which aim to fit \textit{extended} and \textit{enclosed} annotations. We find that the main type of failure case comes from the large opening region, and there is no obvious room boundary to separate the different rooms. To address this difficult scenario, we believe there is a need for more diverse bi-layout training data to ensure our model can learn the corresponding label properties.

\section{Future Directions}
\label{supp_future_work}
We provide possible future research directions based on our current proposed method.

\noindent \textbf{Cross-dataset training.}
Our main manuscript shows that pretraining on the large-scale ZInD~\cite{cruz2021zillow} dataset can benefit the model performance evaluated on the MatterportLayout~\cite{zou2021manhattan} dataset. This observation provides the possible direction for cross-dataset training, which may further improve the model performance.

\noindent \textbf{Bi{-}Layout to multiple layouts.}
Our Bi{-}Layout model can generate two types of predictions. Based on our network design, it is possible to extend the number of predictions to more than two predictions. In the future, once the dataset provides multiple types of labels, multiple predictions can be achieved by simply adding more global context embeddings and the corresponding prediction heads. Most importantly, due to our shared feature guidance module design, those additional components for multiple predictions are very lightweight, which can still maintain the compactness of our model.





%% file: figs/supp_relabel_flow.tex
\begin{figure}[tp]
    \centering
    \includegraphics[width=\linewidth]{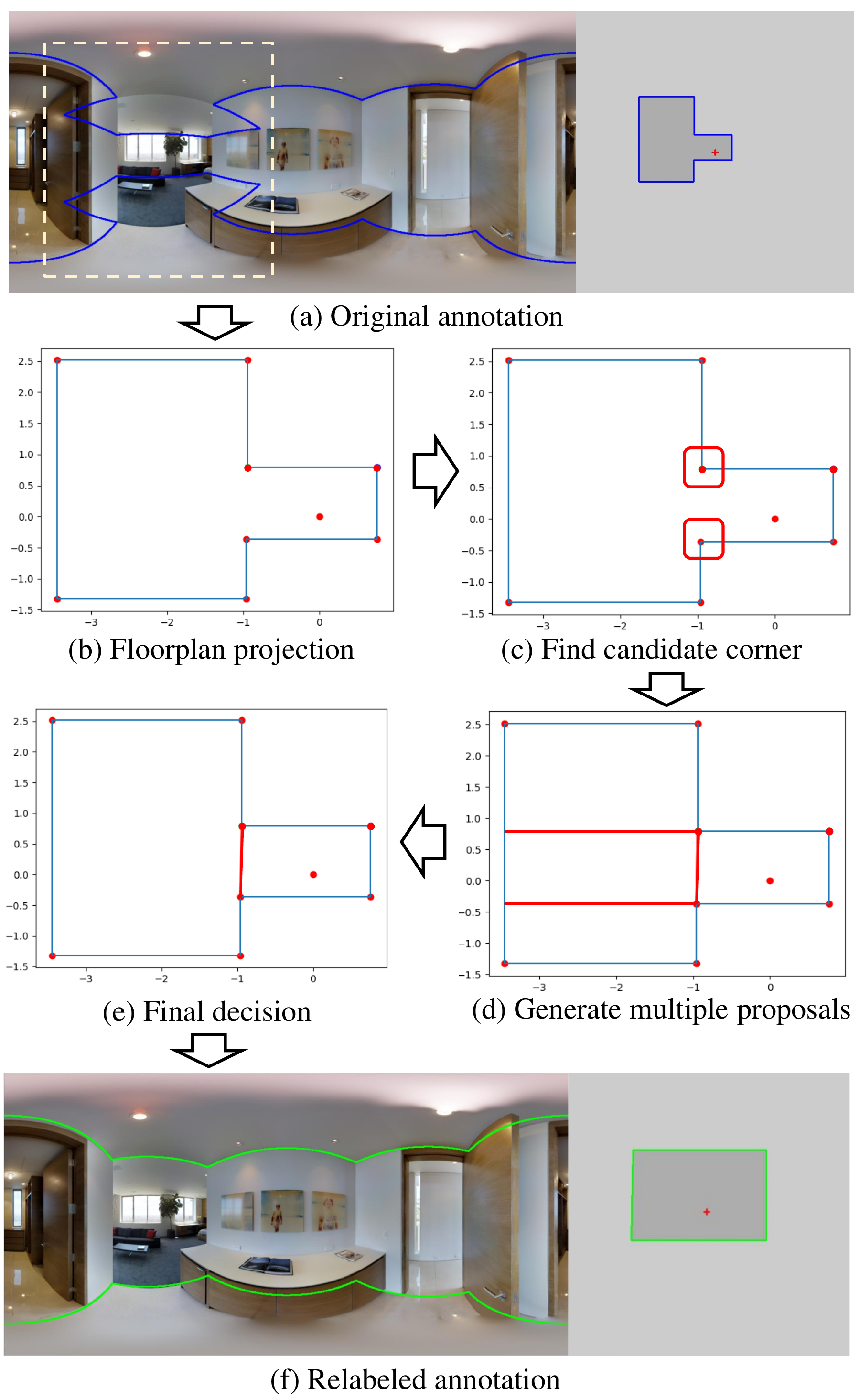}
    \vspace{-4mm}
    \caption{\textbf{Our relabeling pipeline}. \blue{Blue line} in (a) and \green{Green line} in (f) represent the \textbf{original annotation} and \textbf{our relabeled annotation}, respectively. 
    The layout boundaries are shown on the left, and their bird's-eye view projections are on the right. The dashed lines in (a) highlight the occlusion region in the original label.}
    \label{fig:supp_relabel_flow}
    \vspace{-6mm}
\end{figure}

%% file: figs/supp_relabel_results.tex
\begin{figure*}[tp]
    \centering
    \includegraphics[width=0.9\linewidth]{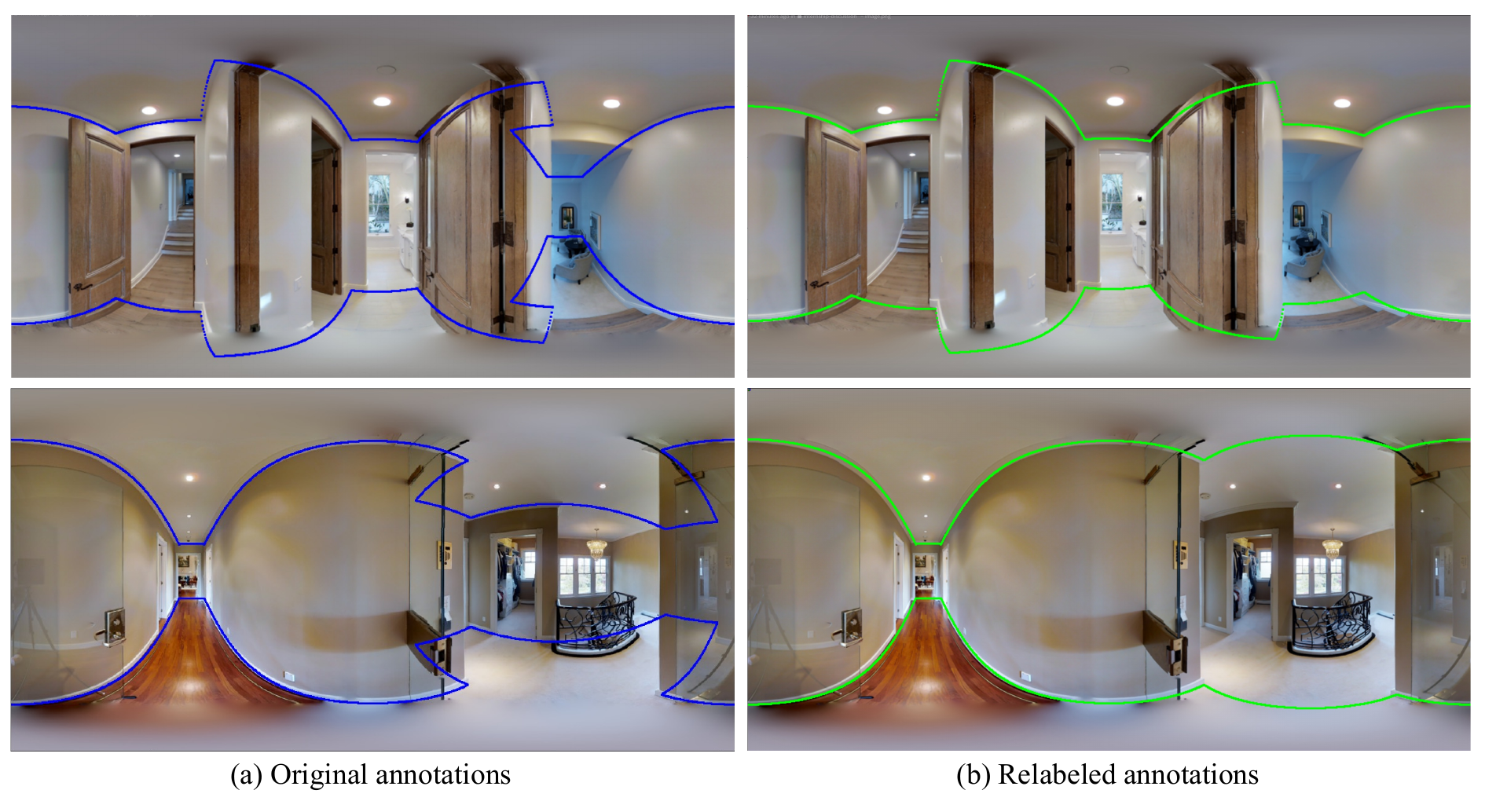}
    \caption{\textbf{Bi-layout annotations on the MatterportLayout~\cite{zou2021manhattan} dataset.} \blue{Blue} and \green{Green} lines indicate original and relabeled annotations. 
    }
    \label{fig:supp_relabel_results}
\end{figure*}

%% file: figs/supp_zind_gt.tex
\begin{figure*}[tp]
    \centering
    \includegraphics[width=0.9\linewidth]{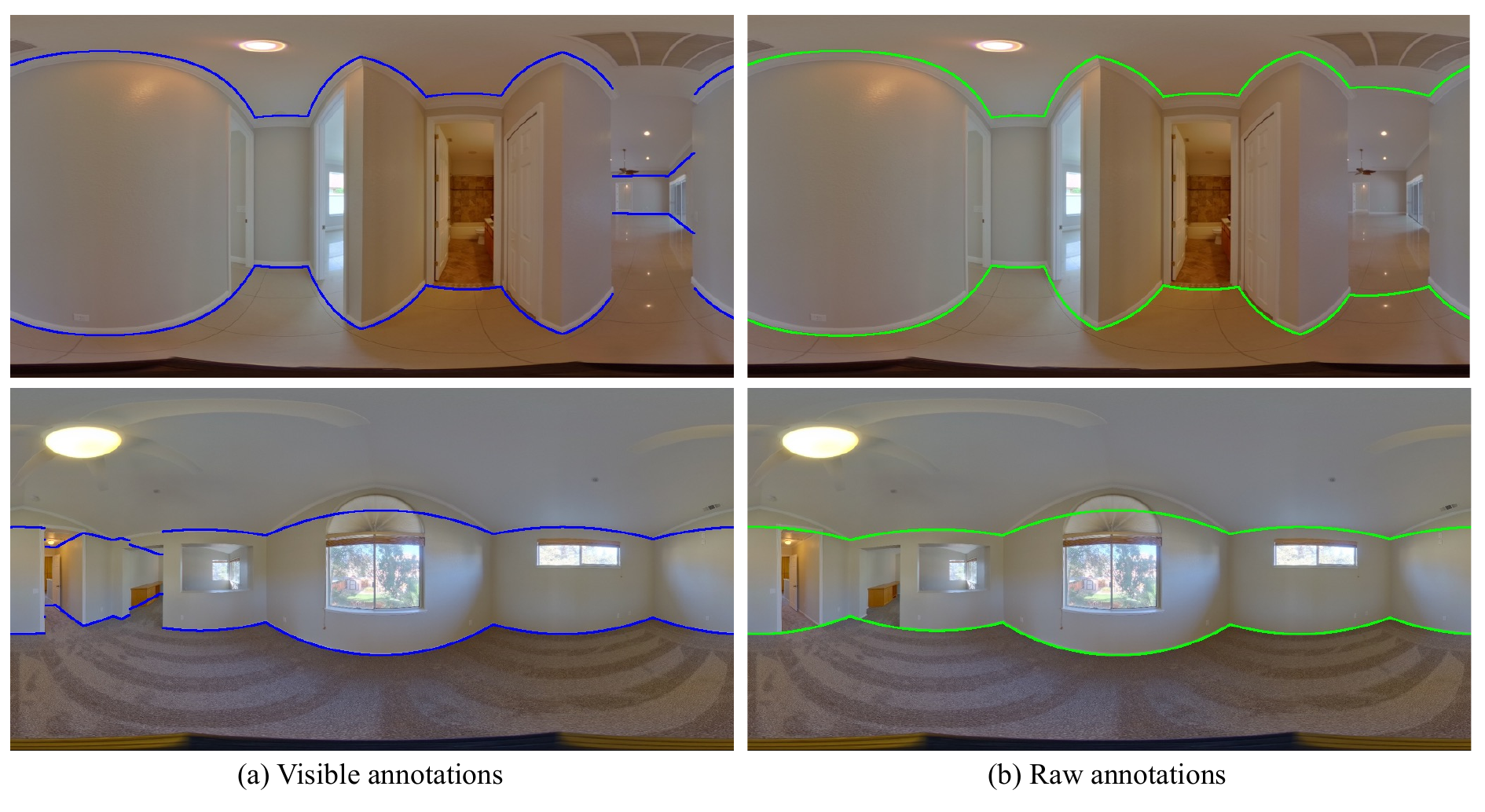}
    \caption{\textbf{Bi-layout annotations on the ZInD~\cite{cruz2021zillow} dataset.} \blue{Blue} and \green{Green} lines indicate \textit{visible} and \textit{raw} annotations. 
    }
    \label{fig:supp_zind_gt}
\end{figure*}

%% file: figs/mp3d_supp.tex
\begin{figure*}[tp]
  \centering
  \includegraphics[width=0.95\linewidth]{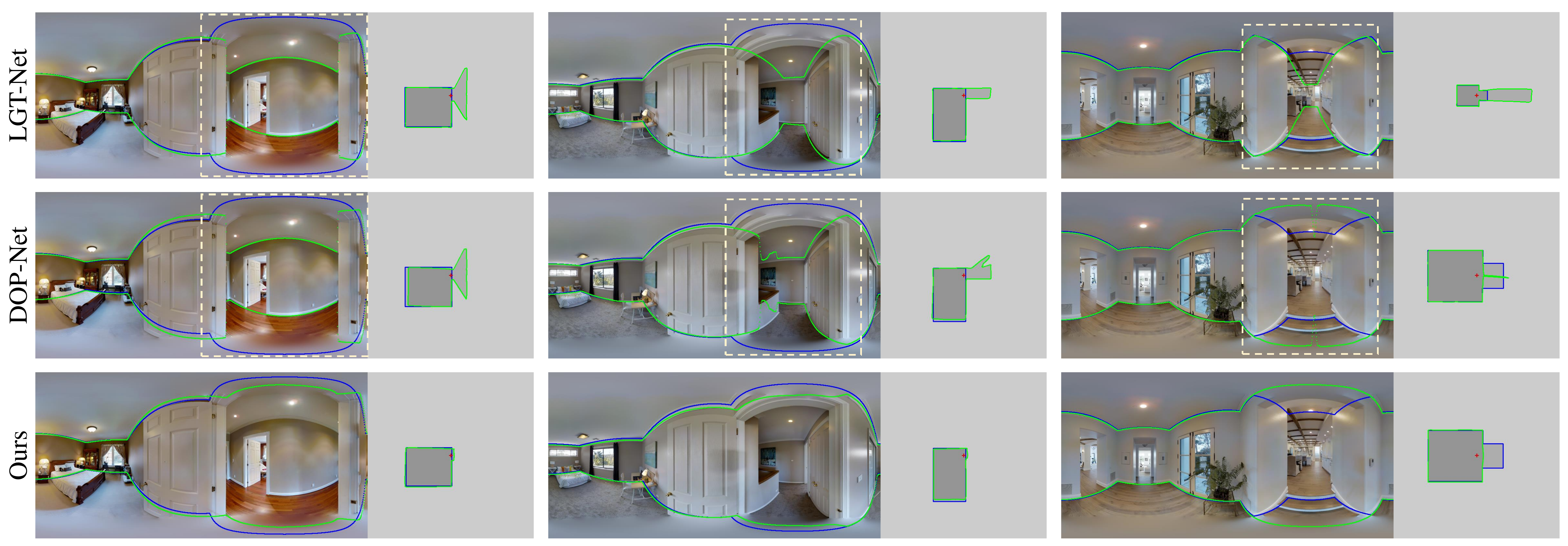}
  \caption{\textbf{More qualitative results on the MatterportLayout~\cite{zou2021manhattan} dataset.} \blue{Blue} and \green{Green} represent ground truth labels and predictions, respectively. The boundaries of the room layout are on the left, and their bird's-eye view projections are on the right. We show our \textit{disambiguate} results, which effectively address the ambiguity issue, while the SoTA methods struggle with the ambiguity, as highlighted in dashed lines.
  } 
  \label{fig:mp3d_supp}
\end{figure*}

%% file: figs/zind_supp.tex
\begin{figure*}[tp]
  \centering
  \includegraphics[width=0.95\linewidth]{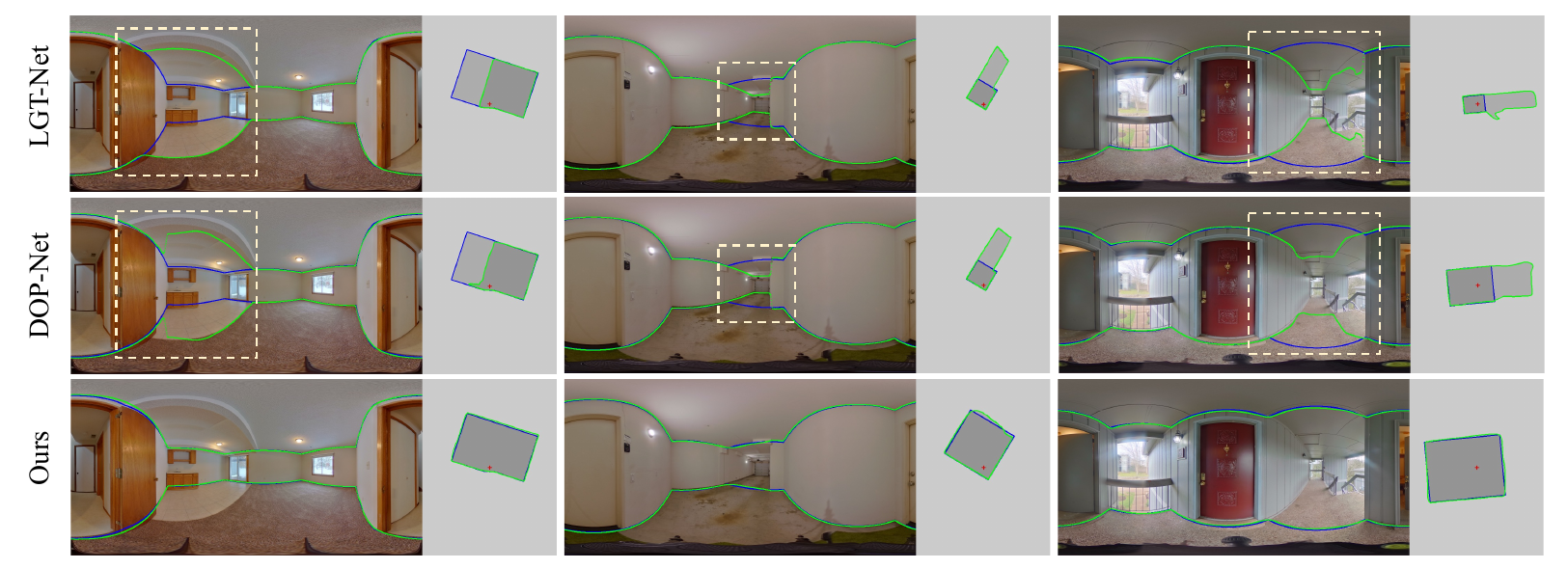}
  \caption{\textbf{More qualitative results on the ZInD~\cite{cruz2021zillow} dataset.} \blue{Blue} and \green{Green} represent ground truth labels and predictions, respectively. The boundaries of the room layout are on the left, and their bird's-eye view projections are on the right. We show our \textit{disambiguate} results, which effectively address the ambiguity issue, while the SoTA methods struggle with the ambiguity, as highlighted in dashed lines.
  } 
  \label{fig:zind_supp}
\end{figure*}

%% file: figs/supp_ambiguity_detection.tex
\begin{figure*}[tp]
  \centering
  \includegraphics[width=0.95\linewidth]{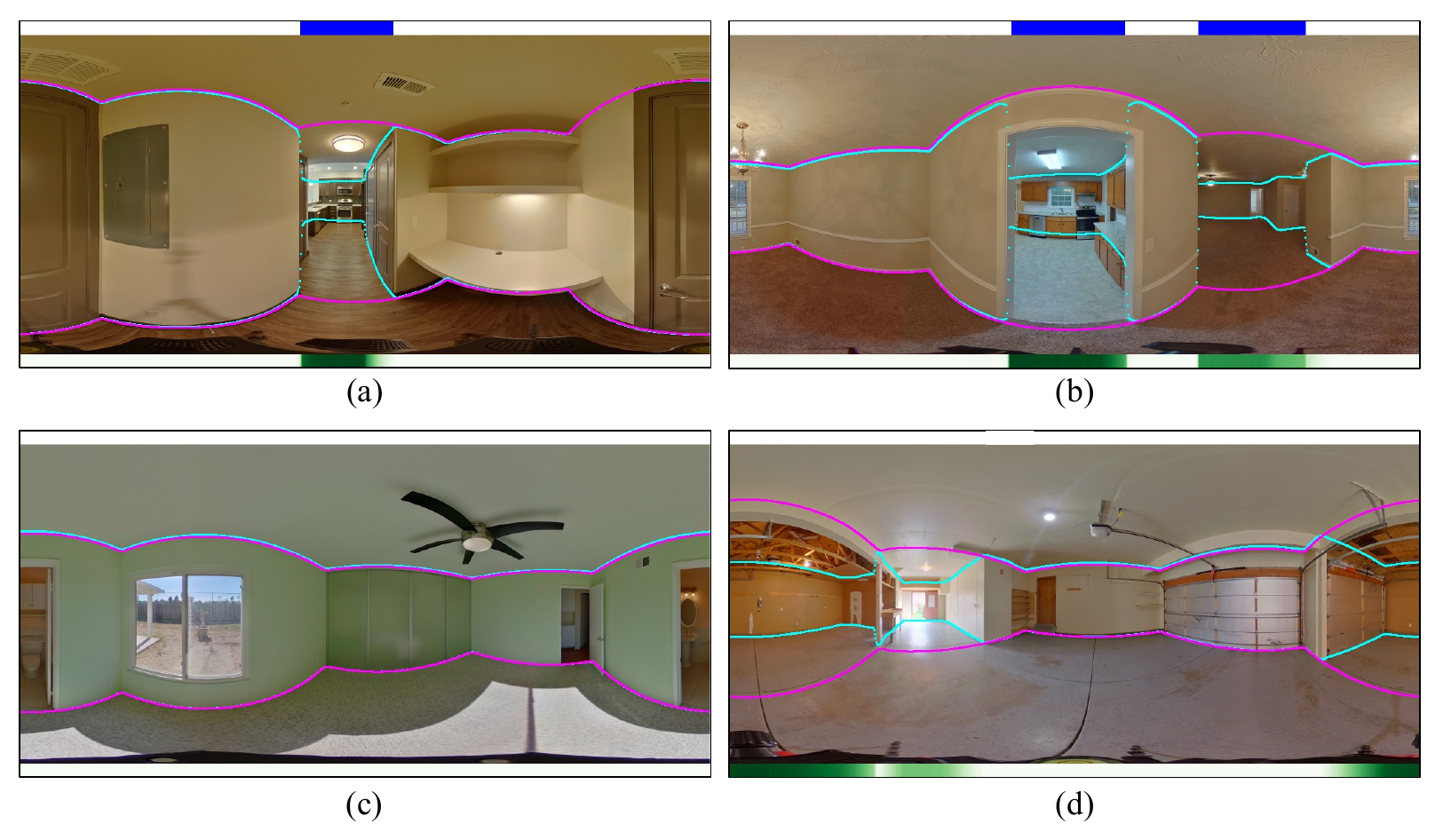}
  \caption{\textbf{Qualitative results and different scenarios for ambiguity detection.} \blue{Blue} and \green{Green} on the top and bottom rows per image represent ground truth and predicted confidence, respectively. \textcolor{cyan}{Cyan} and \textcolor{magenta}{Magenta} lines are our \textit{extended} and \textit{enclosed} type layout predictions. 
  In (a) and (b), our Bi{-}Layout model can accurately detect ambiguous regions as the GT shows. In (c), our model is able to predict two identical predictions when there is no ambiguous region in the image. In (d), we show a special case where the GT does not indicate the ambiguous regions as it should be (i.e., GT itself has ambiguity), and our model can still successfully identify them. Note that (d) is the example we show in the main manuscript where the SoTA methods fail, which corroborates the ambiguity in this image.
  } 
  \label{fig:supp_ambiguity_detection}
\end{figure*}

%% file: tables/supp_single_comparison.tex
\begin{table*}[tp]
\centering
  \resizebox{0.8\linewidth}{!} 
  {
  \centering
    \begin{tabular}{lccccc}
    \toprule
    & & \multicolumn{2}{c}{Full set} & \multicolumn{2}{c}{Subset}  \\
    \midrule
    Method & \# Params & 2DIoU(\%) & 3DIoU(\%) & 2DIoU(\%) & 3DIoU(\%) \\
    \midrule
    LGT-Net~\cite{jiang2022lgt} & 136 M & 83.52 & 81.11 & 53.17 & 50.54 \\
    \rowcolor{ourcolor}
    Our single branch & 102 M  & \textbf{84.09} & \textbf{81.78}  & \textbf{58.65} & \textbf{56.23} \\
    \bottomrule
    \end{tabular}%
  }
  \caption{\textbf{Global Context Embedding for a single branch.} We conduct the experiment on both the full set and subset of the MatterportLayout~\cite{zou2021manhattan} dataset. We choose LGT-Net~\cite{jiang2022lgt} as our baseline method to compare the effectiveness of our Global Context Embedding design.}
  \label{tab:embedding_supp}%
\end{table*}%

%% file: tables/supp_model_variation.tex
\begin{table*}[tp]
\centering
  \resizebox{0.8\linewidth}{!} 
  {
  \centering
    \begin{tabular}{lccccc}
    \toprule
    & & \multicolumn{2}{c}{Full set} & \multicolumn{2}{c}{Subset}  \\
    \midrule
    Method & \# Params & 2DIoU(\%) & 3DIoU(\%) & 2DIoU(\%) & 3DIoU(\%) \\
    \midrule
    Two models & 272 M  & \textbf{85.29} & \textbf{82.72} & 62.54 & \textbf{60.04} \\
    Two transformers & 203 M  & 84.35 & 81.88 & 59.21 & 56.80\\
    Two heads & 136 M  & 84.06 & 81.51 & 57.97 & 55.47\\
    \midrule
    \rowcolor{ourcolor}
    Ours (c = 512) & 102 M  &  85.10  &  82.57 & \textbf{62.81} & 59.97 \\
    \bottomrule
    \end{tabular}%
  }
  \caption{\textbf{Model size and performance trade-off.}
  We conduct the experiment on both the full set and subset of the MatterportLayout~\cite{zou2021manhattan} dataset and evaluate with our proposed \textit{disambiguate} metric.
  }
  \label{tab:model_size_supp}%
\end{table*}%

%% file: tables/supp_parameter.tex
\begin{table*}[tp]
\centering
  \resizebox{0.8\linewidth}{!} 
  {
  \centering
    \begin{tabular}{lccccc}
    \toprule
    & & \multicolumn{2}{c}{Full set} & \multicolumn{2}{c}{Subset}  \\
    \midrule
    Method & \# Params & 2DIoU(\%) & 3DIoU(\%) & 2DIoU(\%) & 3DIoU(\%) \\
    \midrule
    Ours (c = 1024) & 172 M  & \textbf{85.25} & \textbf{82.76} & \textbf{63.33} & \textbf{60.50} \\
    Ours (c = 512) & 102 M  &  85.10  &  82.57 & 62.81 & 59.97 \\
    Ours (c = 256) & 80 M  &  84.47  &  81.90 & 60.39 & 57.83\\
    \bottomrule
    \end{tabular}%
  }
  \caption{\textbf{Different image feature dimensions.}
  We conduct the experiment on the full set and subset of the MatterportLayout~\cite{zou2021manhattan} dataset and evaluate with our proposed \textit{disambiguate} metric.
  }
  \label{tab:our_diff_model_supp}%
\end{table*}%

%% file: tables/pretraining.tex
\begin{table}[tp]
  \centering
    \begin{tabular}{lcc}
    \toprule
    Model & 2D IoU (\%) & 3D IoU (\%) \\
    \midrule
    Train from scratch & 85.10 & 82.57 \\
    Pretrain on ZInD-Simple & 85.52  & 83.28 \\
    \rowcolor{ourcolor} 
    Pretrain on ZInD-All & \textbf{85.81} & \textbf{83.52} \\
    \bottomrule
    \end{tabular}%
  \caption{\textbf{Pretraining effectiveness} on MatterportLayout~\cite{zou2021manhattan} with our proposed \textit{disambiguate} metric. The pretraining on different types of ZInD~\cite{cruz2021zillow} datasets indeed helps the model to disambiguate, and the more data for the pretraining stage, the more performance gain it has.}
  \label{tab:pretraining}%
\end{table}%

%% file: figs/supp_failure_case.tex
\begin{figure}[tp]
    \centering
    \includegraphics[width=\linewidth]{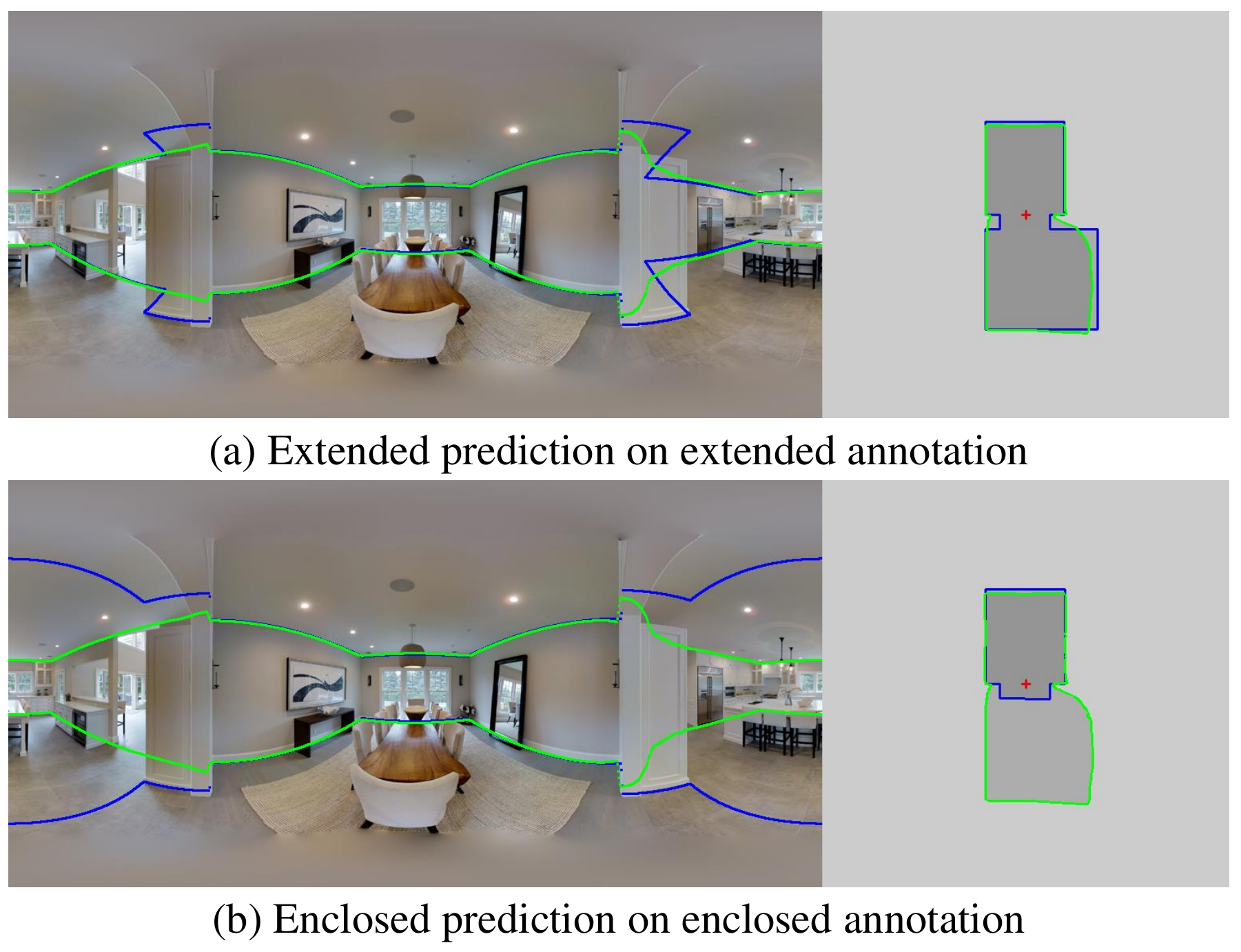}
    \caption{\textbf{Failure case on the MatterportLayout~\cite{zou2021manhattan} dataset.} \blue{Blue} and \green{Green} represent ground truth labels and predictions, respectively. The boundaries of the room layout are on the left, and their bird's eye view projections are on the right.}
    \label{fig:supp_failure_case}
\end{figure}

%% file: _main.bbl
\begin{thebibliography}{45}
\providecommand{\natexlab}[1]{#1}
\providecommand{\url}[1]{\texttt{#1}}
\expandafter\ifx\csname urlstyle\endcsname\relax
  \providecommand{\doi}[1]{doi: #1}\else
  \providecommand{\doi}{doi: \begingroup \urlstyle{rm}\Url}\fi

\bibitem[Bello et~al.(2019)Bello, Zoph, Vaswani, Shlens, and Le]{bello2019attention}
Irwan Bello, Barret Zoph, Ashish Vaswani, Jonathon Shlens, and Quoc~V Le.
\newblock Attention augmented convolutional networks.
\newblock In \emph{ICCV}, 2019.

\bibitem[Carion et~al.(2020)Carion, Massa, Synnaeve, Usunier, Kirillov, and Zagoruyko]{carion2020end}
Nicolas Carion, Francisco Massa, Gabriel Synnaeve, Nicolas Usunier, Alexander Kirillov, and Sergey Zagoruyko.
\newblock End-to-end object detection with transformers.
\newblock In \emph{ECCV}, 2020.

\bibitem[Chen et~al.(2022)Chen, Zhao, Zhou, and Zhang]{chen2022pq}
Xiaoxue Chen, Hao Zhao, Guyue Zhou, and Ya-Qin Zhang.
\newblock Pq-transformer: Jointly parsing 3d objects and layouts from point clouds.
\newblock \emph{IEEE Robotics and Automation Letters}, 2022.

\bibitem[Cheng et~al.(2021)Cheng, Schwing, and Kirillov]{cheng2021per}
Bowen Cheng, Alex Schwing, and Alexander Kirillov.
\newblock Per-pixel classification is not all you need for semantic segmentation.
\newblock \emph{NeurIPS}, 2021.

\bibitem[Cheng et~al.(2022)Cheng, Misra, Schwing, Kirillov, and Girdhar]{cheng2022masked}
Bowen Cheng, Ishan Misra, Alexander~G Schwing, Alexander Kirillov, and Rohit Girdhar.
\newblock Masked-attention mask transformer for universal image segmentation.
\newblock In \emph{CVPR}, 2022.

\bibitem[Coughlan and Yuille(1999)]{coughlan1999manhattan}
James~M Coughlan and Alan~L Yuille.
\newblock Manhattan world: Compass direction from a single image by bayesian inference.
\newblock In \emph{ICCV}, 1999.

\bibitem[Cruz et~al.(2021)Cruz, Hutchcroft, Li, Khosravan, Boyadzhiev, and Kang]{cruz2021zillow}
Steve Cruz, Will Hutchcroft, Yuguang Li, Naji Khosravan, Ivaylo Boyadzhiev, and Sing~Bing Kang.
\newblock Zillow indoor dataset: Annotated floor plans with 360deg panoramas and 3d room layouts.
\newblock In \emph{CVPR}, 2021.

\bibitem[Fernandez-Labrador et~al.(2020)Fernandez-Labrador, Facil, Perez-Yus, Demonceaux, Civera, and Guerrero]{fernandez2020corners}
Clara Fernandez-Labrador, Jose~M Facil, Alejandro Perez-Yus, C{\'e}dric Demonceaux, Javier Civera, and Jose~J Guerrero.
\newblock Corners for layout: End-to-end layout recovery from 360 images.
\newblock \emph{IEEE Robotics and Automation Letters}, 2020.

\bibitem[Greene(1986)]{greene1986environment}
Ned Greene.
\newblock Environment mapping and other applications of world projections.
\newblock \emph{IEEE Computer Graphics and Applications}, 1986.

\bibitem[Gupta et~al.(2010)Gupta, Hebert, Kanade, and Blei]{gupta2010estimating}
Abhinav Gupta, Martial Hebert, Takeo Kanade, and David Blei.
\newblock Estimating spatial layout of rooms using volumetric reasoning about objects and surfaces.
\newblock \emph{NeurIPS}, 2010.

\bibitem[He et~al.(2016)He, Zhang, Ren, and Sun]{he2016deep}
Kaiming He, Xiangyu Zhang, Shaoqing Ren, and Jian Sun.
\newblock Deep residual learning for image recognition.
\newblock In \emph{CVPR}, 2016.

\bibitem[Hedau et~al.(2009)Hedau, Hoiem, and Forsyth]{hedau2009recovering}
Varsha Hedau, Derek Hoiem, and David Forsyth.
\newblock Recovering the spatial layout of cluttered rooms.
\newblock In \emph{ICCV}, 2009.

\bibitem[Hirzer et~al.(2020)Hirzer, Lepetit, and ROTH]{hirzer2020smart}
Martin Hirzer, Vincent Lepetit, and PETER ROTH.
\newblock Smart hypothesis generation for efficient and robust room layout estimation.
\newblock In \emph{WACV}, 2020.

\bibitem[Hochreiter and Schmidhuber(1997)]{hochreiter1997long}
Sepp Hochreiter and J{\"u}rgen Schmidhuber.
\newblock Long short-term memory.
\newblock \emph{Neural Computation}, 1997.

\bibitem[Huang et~al.(2018)Huang, Qi, Zhu, Xiao, Xu, and Zhu]{huang2018holistic}
Siyuan Huang, Siyuan Qi, Yixin Zhu, Yinxue Xiao, Yuanlu Xu, and Song-Chun Zhu.
\newblock Holistic 3d scene parsing and reconstruction from a single rgb image.
\newblock In \emph{ECCV}, 2018.

\bibitem[Huang and Belongie(2017)]{huang2017arbitrary}
Xun Huang and Serge Belongie.
\newblock Arbitrary style transfer in real-time with adaptive instance normalization.
\newblock In \emph{ICCV}, 2017.

\bibitem[Jiang et~al.(2022)Jiang, Xiang, Xu, and Zhao]{jiang2022lgt}
Zhigang Jiang, Zhongzheng Xiang, Jinhua Xu, and Ming Zhao.
\newblock Lgt-net: Indoor panoramic room layout estimation with geometry-aware transformer network.
\newblock In \emph{CVPR}, 2022.

\bibitem[Kipf and Welling(2017)]{kipf2017semi}
Thomas~N Kipf and Max Welling.
\newblock Semi-supervised classification with graph convolutional networks.
\newblock In \emph{ICLR}, 2017.

\bibitem[Mallya and Lazebnik(2015)]{mallya2015learning}
Arun Mallya and Svetlana Lazebnik.
\newblock Learning informative edge maps for indoor scene layout prediction.
\newblock In \emph{ICCV}, 2015.

\bibitem[Meinhardt et~al.(2022)Meinhardt, Kirillov, Leal-Taixe, and Feichtenhofer]{meinhardt2022trackformer}
Tim Meinhardt, Alexander Kirillov, Laura Leal-Taixe, and Christoph Feichtenhofer.
\newblock Trackformer: Multi-object tracking with transformers.
\newblock In \emph{CVPR}, 2022.

\bibitem[Parmar et~al.(2018)Parmar, Vaswani, Uszkoreit, Kaiser, Shazeer, Ku, and Tran]{parmar2018image}
Niki Parmar, Ashish Vaswani, Jakob Uszkoreit, Lukasz Kaiser, Noam Shazeer, Alexander Ku, and Dustin Tran.
\newblock Image transformer.
\newblock In \emph{ICML}, 2018.

\bibitem[Paszke et~al.(2019)Paszke, Gross, Massa, Lerer, Bradbury, Chanan, Killeen, Lin, Gimelshein, Antiga, et~al.]{paszke2019pytorch}
Adam Paszke, Sam Gross, Francisco Massa, Adam Lerer, James Bradbury, Gregory Chanan, Trevor Killeen, Zeming Lin, Natalia Gimelshein, Luca Antiga, et~al.
\newblock Pytorch: An imperative style, high-performance deep learning library.
\newblock \emph{NeurIPS}, 2019.

\bibitem[Perez et~al.(2018)Perez, Strub, De~Vries, Dumoulin, and Courville]{perez2018film}
Ethan Perez, Florian Strub, Harm De~Vries, Vincent Dumoulin, and Aaron Courville.
\newblock Film: Visual reasoning with a general conditioning layer.
\newblock In \emph{AAAI}, 2018.

\bibitem[Pintore et~al.(2020)Pintore, Agus, and Gobbetti]{pintore2020atlantanet}
Giovanni Pintore, Marco Agus, and Enrico Gobbetti.
\newblock Atlantanet: inferring the 3d indoor layout from a single 360∘ image beyond the manhattan world assumption.
\newblock In \emph{ECCV}, 2020.

\bibitem[Ramalingam et~al.(2013)Ramalingam, Pillai, Jain, and Taguchi]{ramalingam2013manhattan}
Srikumar Ramalingam, Jaishanker~K Pillai, Arpit Jain, and Yuichi Taguchi.
\newblock Manhattan junction catalogue for spatial reasoning of indoor scenes.
\newblock In \emph{CVPR}, 2013.

\bibitem[Ren et~al.(2017)Ren, Li, Chen, and Kuo]{ren2017coarse}
Yuzhuo Ren, Shangwen Li, Chen Chen, and C-C~Jay Kuo.
\newblock A coarse-to-fine indoor layout estimation (cfile) method.
\newblock In \emph{ACCV}, 2017.

\bibitem[Schuster and Paliwal(1997)]{schuster1997bidirectional}
Mike Schuster and Kuldip~K Paliwal.
\newblock Bidirectional recurrent neural networks.
\newblock \emph{IEEE Transactions on Signal Processing}, 1997.

\bibitem[Schwing and Urtasun(2012)]{schwing2012efficient}
Alexander~G Schwing and Raquel Urtasun.
\newblock Efficient exact inference for 3d indoor scene understanding.
\newblock In \emph{ECCV}, 2012.

\bibitem[Shen et~al.(2023)Shen, Zheng, Lin, Nie, Liao, Zheng, and Zhao]{shen2023dopnet}
Zhijie Shen, Zishuo Zheng, Chunyu Lin, Lang Nie, Kang Liao, Shuai Zheng, and Yao Zhao.
\newblock Disentangling orthogonal planes for indoor panoramic room layout estimation with cross-scale distortion awareness.
\newblock In \emph{CVPR}, 2023.

\bibitem[Su et~al.(2023)Su, Tung, Peng, Wonka, and Chu]{su2023slibo}
Jheng-Wei Su, Kuei-Yu Tung, Chi-Han Peng, Peter Wonka, and Hung-Kuo Chu.
\newblock Slibo-net: Floorplan reconstruction via slicing box representation with local geometry regularization.
\newblock In \emph{NeurIPS}, 2023.

\bibitem[Sun et~al.(2019)Sun, Hsiao, Sun, and Chen]{sun2019horizonnet}
Cheng Sun, Chi-Wei Hsiao, Min Sun, and Hwann-Tzong Chen.
\newblock Horizonnet: Learning room layout with 1d representation and pano stretch data augmentation.
\newblock In \emph{CVPR}, 2019.

\bibitem[Sun et~al.(2021)Sun, Sun, and Chen]{sun2021hohonet}
Cheng Sun, Min Sun, and Hwann-Tzong Chen.
\newblock Hohonet: 360 indoor holistic understanding with latent horizontal features.
\newblock In \emph{CVPR}, 2021.

\bibitem[Vaswani et~al.(2017)Vaswani, Shazeer, Parmar, Uszkoreit, Jones, Gomez, Kaiser, and Polosukhin]{vaswani2017attention}
Ashish Vaswani, Noam Shazeer, Niki Parmar, Jakob Uszkoreit, Llion Jones, Aidan~N Gomez, {\L}ukasz Kaiser, and Illia Polosukhin.
\newblock Attention is all you need.
\newblock \emph{NeurIPS}, 2017.

\bibitem[Wang et~al.(2021)Wang, Yeh, Sun, Chiu, and Tsai]{wang2021led2}
Fu-En Wang, Yu-Hsuan Yeh, Min Sun, Wei-Chen Chiu, and Yi-Hsuan Tsai.
\newblock Led2-net: Monocular 360deg layout estimation via differentiable depth rendering.
\newblock In \emph{CVPR}, 2021.

\bibitem[Wang et~al.(2013)Wang, Gould, and Roller]{wang2013discriminative}
Huayan Wang, Stephen Gould, and Daphne Roller.
\newblock Discriminative learning with latent variables for cluttered indoor scene understanding.
\newblock \emph{Communications of the ACM}, 2013.

\bibitem[Yang et~al.(2019)Yang, Wang, Peng, Wonka, Sun, and Chu]{yang2019dula}
Shang-Ta Yang, Fu-En Wang, Chi-Han Peng, Peter Wonka, Min Sun, and Hung-Kuo Chu.
\newblock Dula-net: A dual-projection network for estimating room layouts from a single rgb panorama.
\newblock In \emph{CVPR}, 2019.

\bibitem[Yue et~al.(2023)Yue, Kontogianni, Schindler, and Engelmann]{yue2023connecting}
Yuanwen Yue, Theodora Kontogianni, Konrad Schindler, and Francis Engelmann.
\newblock Connecting the dots: Floorplan reconstruction using two-level queries.
\newblock In \emph{CVPR}, 2023.

\bibitem[Zhang et~al.(2021)Zhang, Cui, Zhang, Zeng, Pollefeys, and Liu]{zhang2021holistic}
Cheng Zhang, Zhaopeng Cui, Yinda Zhang, Bing Zeng, Marc Pollefeys, and Shuaicheng Liu.
\newblock Holistic 3d scene understanding from a single image with implicit representation.
\newblock In \emph{CVPR}, 2021.

\bibitem[Zhao et~al.(2017)Zhao, Lu, Yao, Guo, Chen, and Zhang]{zhao2017physics}
Hao Zhao, Ming Lu, Anbang Yao, Yiwen Guo, Yurong Chen, and Li Zhang.
\newblock Physics inspired optimization on semantic transfer features: An alternative method for room layout estimation.
\newblock In \emph{CVPR}, 2017.

\bibitem[Zhao et~al.(2021)Zhao, Han, Zhang, Xu, and Cheng]{zhao2021deep}
Kai Zhao, Qi Han, Chang-Bin Zhang, Jun Xu, and Ming-Ming Cheng.
\newblock Deep hough transform for semantic line detection.
\newblock \emph{TPAMI}, 2021.

\bibitem[Zhao et~al.(2022)Zhao, Wen, Xue, and Gao]{zhao20223d}
Yining Zhao, Chao Wen, Zhou Xue, and Yue Gao.
\newblock 3d room layout estimation from a cubemap of panorama image via deep manhattan hough transform.
\newblock In \emph{ECCV}, 2022.

\bibitem[Zhou et~al.(2022)Zhou, Yin, Koltun, and Kr{\"a}henb{\"u}hl]{zhou2022global}
Xingyi Zhou, Tianwei Yin, Vladlen Koltun, and Philipp Kr{\"a}henb{\"u}hl.
\newblock Global tracking transformers.
\newblock In \emph{CVPR}, 2022.

\bibitem[Zhu et~al.(2021)Zhu, Su, Lu, Li, Wang, and Dai]{zhu2021deformable}
Xizhou Zhu, Weijie Su, Lewei Lu, Bin Li, Xiaogang Wang, and Jifeng Dai.
\newblock Deformable detr: Deformable transformers for end-to-end object detection.
\newblock In \emph{ICLR}, 2021.

\bibitem[Zou et~al.(2018)Zou, Colburn, Shan, and Hoiem]{zou2018layoutnet}
Chuhang Zou, Alex Colburn, Qi Shan, and Derek Hoiem.
\newblock Layoutnet: Reconstructing the 3d room layout from a single rgb image.
\newblock In \emph{CVPR}, 2018.

\bibitem[Zou et~al.(2021)Zou, Su, Peng, Colburn, Shan, Wonka, Chu, and Hoiem]{zou2021manhattan}
Chuhang Zou, Jheng-Wei Su, Chi-Han Peng, Alex Colburn, Qi Shan, Peter Wonka, Hung-Kuo Chu, and Derek Hoiem.
\newblock Manhattan room layout reconstruction from a single 360\^{} ∘ 360∘ image: A comparative study of state-of-the-art methods.
\newblock \emph{IJCV}, 2021.

\end{thebibliography}
